%% file: main.tex
\documentclass[10pt,journal,compsoc]{IEEEtran}

\usepackage{graphicx}
\usepackage{booktabs} 
\usepackage{float}
\usepackage{gensymb}
\usepackage{tabularx}
\usepackage{tikz}
\usepackage{comment}
\usepackage{amsmath,amssymb} 
\usepackage{color}
\usepackage{amsfonts}
\usepackage{caption}
\usepackage{subfigure}
\usepackage{multirow}
\usepackage{mathtools}
\usepackage{tablefootnote}
\usepackage{dsfont}
\usepackage{ragged2e}
\usepackage[pagebackref,breaklinks,colorlinks,urlcolor=black,citecolor=darkgreen]{hyperref}

\definecolor{darkgreen}{rgb}{0.13, 0.55, 0.13}
\newcommand{\OURS}{RIGA}
\newcommand{\defeq}{\vcentcolon=}

\newsavebox\CBox
\def\textBF#1{\sbox\CBox{#1}\resizebox{\wd\CBox}{\ht\CBox}{\textbf{#1}}}
\begin{document}

%
\title{\OURS{}: Rotation-Invariant and Globally-Aware Descriptors for Point Cloud Registration}
%
%
%
%

\author{Hao~Yu,
        Ji~Hou,
        Zheng~Qin,
        Mahdi~Saleh,
        Ivan~Shugurov,
        Kai~Wang,
        Benjamin~Busam,
        Slobodan~Ilic

\IEEEcompsocitemizethanks{\IEEEcompsocthanksitem H.Yu, M.Saleh, I.Shugurov, B.Busam and S.Ilic are with Technical University of Munich, Germany. J.Hou is with Meta Reality Labs, USA. Z.Qin is with National University of Defense Technology, China. K.Wang is with Shanghai Jiao Tong University, China. I.Shugurov and S.Ilic are also with Siemens AG, Germany.\IEEEcompsocthanksitem
E-mail: hao.yu@tum.de
}

}

%
%

\markboth{IEEE TRANSACTIONS ON PATTERN ANALYSIS AND MACHINE INTELLIGENCEE}%
{Shell \MakeLowercase{\textit{et al.}}: Bare Demo of IEEEtran.cls for Computer Society Journals}
\input{sections/0abstract}
\maketitle

\input{sections/1introduction}

\input{sections/2related_work}

\input{sections/3method}

\input{sections/4result}
\input{sections/5conclusion}

\input{sections/6appendix}


\bibliographystyle{IEEEtran}
\bibliography{IEEEabrv,egbib}

%




\end{document}

%% file: sections/0abstract.tex
\IEEEtitleabstractindextext{%
\begin{abstract}
\justifying
Successful point cloud registration relies on accurate correspondences established upon powerful descriptors. However,
existing neural descriptors either leverage a rotation-variant backbone whose performance declines under large rotations, or encode
local geometry that is less distinctive. To address this issue, we introduce RIGA to learn descriptors that are Rotation-Invariant by design
and Globally-Aware. From the Point Pair Features (PPFs) of sparse local
regions, rotation-invariant local geometry is encoded into geometric descriptors. Global awareness of 3D structures and geometric context
is subsequently incorporated, both in a rotation-invariant fashion. More specifically, 3D structures of the whole frame are first represented
by our global PPF signatures, from which structural descriptors are learned to help geometric descriptors sense the 3D world
beyond local regions. Geometric context from the whole scene is then globally aggregated into
descriptors. Finally, the description of sparse regions is interpolated to dense point descriptors,
from which correspondences are extracted for registration. To validate our approach, we conduct extensive experiments on both object- and
scene-level data. With large rotations, RIGA surpasses the state-of-the-art methods by a margin of 8\degree in terms of the Relative Rotation Error on ModelNet40 and improves the Feature Matching Recall by at least 5 percentage points on 3DLoMatch.
\end{abstract}

\begin{IEEEkeywords}
Point Cloud Registration, Rotation-Invariant Descriptors, Globally-Aware Descriptors, Coarse-to-Fine Correspondences
\end{IEEEkeywords}

}

%% file: sections/1introduction.tex
\IEEEraisesectionheading{\section{Introduction}\label{sec:introduction}}

\label{Sec.1}

\IEEEPARstart{O}{ur} entire world is 3D. Modern depth sensors are able to retrieve distance measures of the environment and represent it as point clouds. Naturally, registering point clouds under different sensor poses, a.k.a. point cloud registration,  plays a crucial role in a wide range of real applications such as scene reconstruction, autonomous driving, and simultaneous localization and mapping (SLAM). Given a pair of partially-overlapping point clouds, point cloud registration aims to recover the relative transformation between them. As the relative transformation can be solved in closed-form or estimated by a robust estimator~\cite{fischler1981random} based on putative correspondences, establishing reliable correspondences becomes the key to successful registration. 

\begin{figure}[t!]
	\centering
	\includegraphics[width=0.95\linewidth]{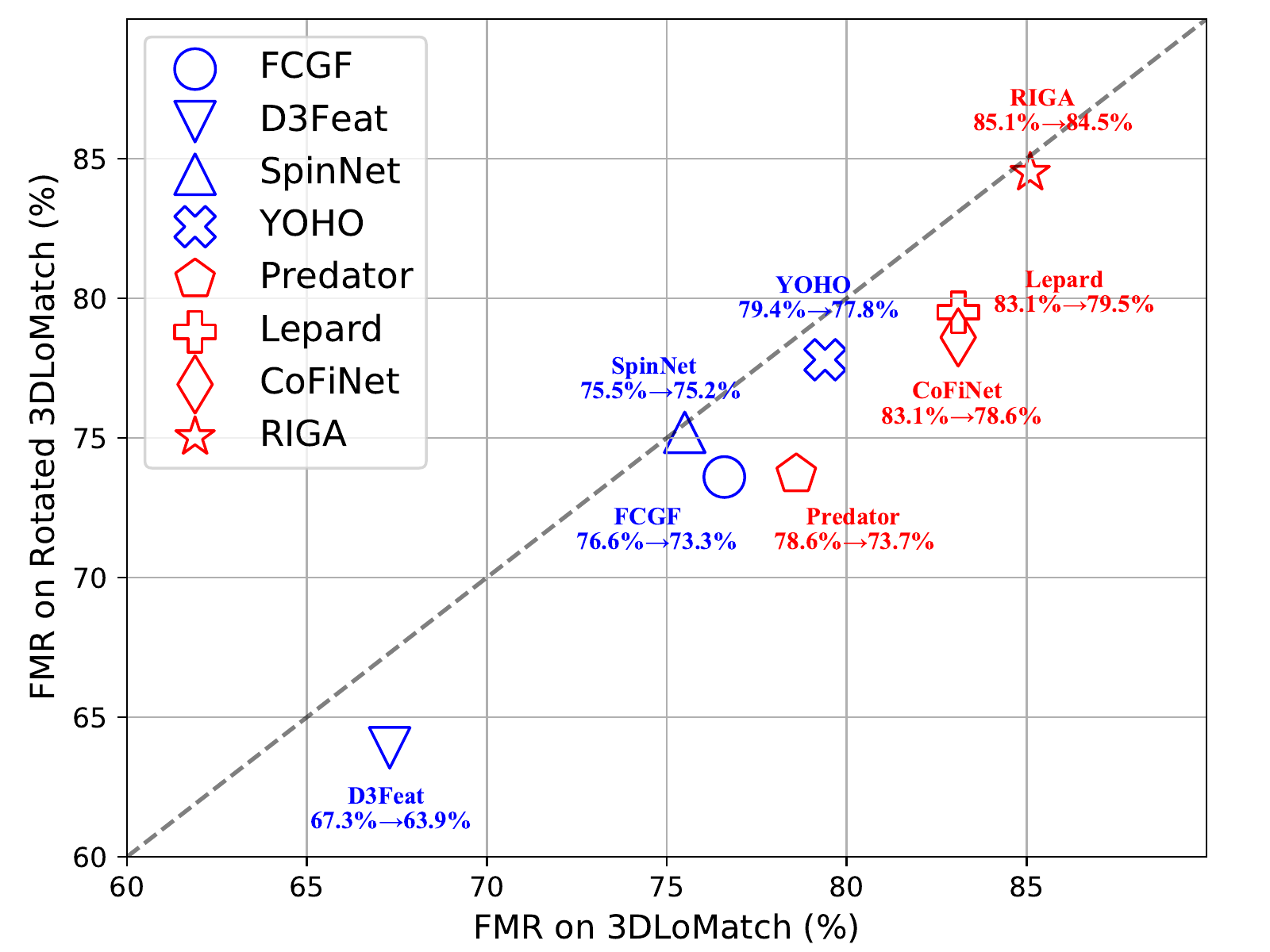}
	\caption{\textbf{Feature Matching Recall (FMR) on 3DLoMatch~\cite{huang2021predator}~(x-axis) and Rotated 3DLoMatch~(y-axis).} Methods that only encode local geometry are marked as blue, while approaches with global awareness are drawn in red. The performance drop from the original~(x-axis) to the rotated~(y-axis) benchmark for each method is also demonstrated. Methods that are more robust against rotations are closer to the 45-degree line. Generally, globally-aware methods perform better on standard benchmarks, while rotation-invariant ones~(SpinNet~\cite{ao2021spinnet}, YOHO~\cite{wang2021you} and \OURS{}) degenerate less under larger rotations. \OURS{} performs the best in both cases with a drop of only 0.6 percent points.}
	\label{fig:fmr}

\end{figure}

\begin{figure*}[h!]
	\centering
	\includegraphics[width=0.95\linewidth]{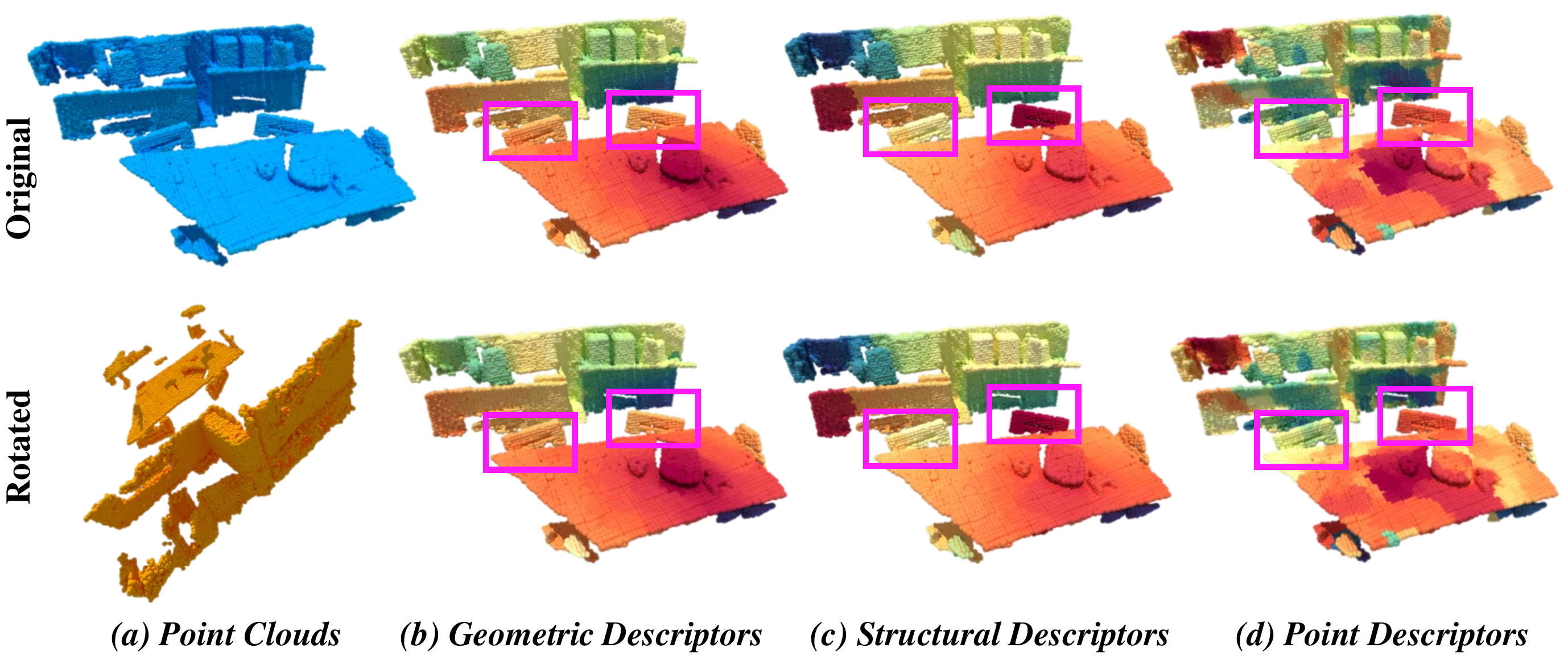}
	\caption{\textbf{Illustration of the Inherent Rotational Invariance and Distinctiveness of \OURS{}.} In (a), an arbitrary rotation is applied to the input scan. \textbf{1) Rotational Invariance}:  In (b), (c) and (d), local, global and point descriptors from \textbf{untrained} \OURS{} are visualized by t-SNE~\cite{van2008visualizing}, respectively. The rotated point cloud is aligned for better visualization. All the descriptors from untrained \OURS{} remain unchanged after rotation~(the second row), which illustrates our inherent rotational invariance guaranteed by design. \textbf{2) Distinctiveness}: In (b), although two chairs inside pink rectangles have similar local geometric descriptors, they are distinguishable in (c) where global structures are encoded, and in (d) where global context is incorporated into local descriptors.}
	\label{fig:teaser}
\end{figure*}

Correspondences are established by matching points according to their associated descriptors. As dense matching is computationally complex, existing works~\cite{deng2018ppfnet,deng2018ppf,gojcic2019perfect,choy2019fully,bai2020d3feat,ao2021spinnet,saleh2020graphite,saleh2022bending,huang2021predator,li2022lepard} widely adopt a first-sampling-then-matching paradigm to match sparse nodes that are either uniformly-sampled or saliently-detected from dense points. Although the computational complexity is significantly reduced, it introduces a new problem of repeatability, i.e., the corresponding points of some nodes are excluded after sparse sampling s.t. they can never be correctly matched. Due to this design, a considerable part of true correspondences is automatically dropped before matching, which significantly constrains the reliability of putative correspondences. To tackle the problem, we have proposed CoFiNet~\cite{yu2021cofinet} which extracts hierarchical correspondences from coarse to fine. On a coarse scale, it learns to match uniformly-sampled nodes whose vicinities share more overlap. The coarse matching significantly shrinks the space of correspondence search of the consecutive stage, where finer correspondences are extracted from the overlapping vicinities. It implicitly considers all the possible correspondences in the matching procedure and therefore eliminates the repeatability issue. However, the descriptors upon which correspondences are extracted by CoFiNet lack robustness against rotations by design. As a consequence, although reliable correspondences are extracted via the proposed coarse-to-fine mechanism,  the performance of CoFiNet still significantly declines when rotations are enlarged, as illustrated in Fig.~\ref{fig:fmr}.

This phenomenon reminds us of the importance of point descriptors and shifts our attention to introducing more powerful descriptors for better registration performance. Recent trends widely adopt neural backbones~\cite{qi2017pointnet, qi2017pointnet++, thomas2019kpconv} to obtain more powerful descriptors~\cite{deng2018ppfnet,deng2018ppf,gojcic2019perfect,choy2019fully,bai2020d3feat,ao2021spinnet,saleh2020graphite,tang2021learning,huang2021predator,wang2021you,yu2021cofinet,li2022lepard,hou2021exploring} from raw points, which gains significant improvement over handcrafted features~\cite{rusu2008aligning,rusu2009fast,drost2010model}. The most recent deep learning-based methods~\cite{ao2021spinnet,huang2021predator,wang2021you,yu2021cofinet,li2022lepard} can be split into two categories according to the way they enhance descriptors. The first one~\cite{ao2021spinnet,wang2021you} aims at promising the rotational invariance of descriptors learned from local geometry by design. For a point $\mathbf{x}_i\in \mathbb{R}^3$ from point cloud $\mathcal{X}$, they propose to guarantee that the local descriptor learned from the support area $\Omega^\mathcal{X}_i$ around $\mathbf{x}_i$ by a model $\mathcal{G}$ is invariant under arbitrary rotations $\mathbf{R}\in \mathit{SO}(3)$, i.e., $\mathcal{G}(\mathbf{R}(\mathbf{x}_i)|\mathbf{R}(\Omega^\mathcal{X}_i)) = \mathcal{G}(\mathbf{x}_i | \Omega^\mathcal{X}_i)$. According to~\cite{deng2018ppf,ao2021spinnet}, these methods are more robust to larger rotations,  which is also demonstrated in Fig.~\ref{fig:fmr}~(see SpinNet, YOHO, and \OURS{}). The second one~\cite{huang2021predator,yu2021cofinet,li2022lepard} instead focuses on incorporating global awareness into local descriptors to enhance the distinctiveness. Compared to descriptors that only encode local geometry, i.e., $\mathcal{G}(\mathbf{x}_i | \Omega^\mathcal{X}_i)$, the globally-aware descriptor $\mathcal{G}(\mathbf{x}_i | \mathcal{X})$ of point $\mathbf{x}_i$ is more distinctive and much easier to be distinguished from other globally-aware descriptors $\mathcal{G}(\mathbf{x}_j | \mathcal{X})$ of points $\mathbf{x}_j$ with $i\neq j$.  As illustrated in Fig.~\ref{fig:teaser}, although it is hard to distinguish two chairs according to local geometry~(Fig.~\ref{fig:teaser}~(b)). However, global awareness helps to separate their description~(Fig.~\ref{fig:teaser}~(c)).
Therefore, globally-aware methods usually perform better on the registration task than approaches that only encode local geometry alone, which is also demonstrated in Fig.~\ref{fig:fmr}. However, each category of methods has its specific drawback -- rotation-invariant descriptors are usually less distinctive due to the blindness to the global context, while globally-aware methods can produce inconsistent descriptions due to the inherent lack of rotational invariance. The current literature lacks an approach that fulfills both aspects simultaneously, i.e., $\mathcal{G}(\mathbf{R}(\mathbf{x}_i)|\mathbf{R}(\mathcal{X})) = \mathcal{G}(\mathbf{x}_i | \mathcal{X})$.


We propose to bridge the lack of globally-aware descriptors that inherently guarantee rotation invariance for the task of point cloud registration with \textbf{\OURS{}}. Our proposed method simultaneously strengthens the robustness against rotations and distinctiveness of learned descriptors, from which coarse-to-fine keypoint-free correspondences are consecutively extracted. More specifically, we adopt a PointNet~\cite{qi2017pointnet} architecture, which takes as input the rotation-invariant handcrafted descriptors to encode rotation-invariant local geometry. To provide a node-specific description of the entire scene in a rotation-invariant fashion, we design global PPF signatures that describe each node by considering the spatial relationship of the remaining nodes w.r.t. it. Subsequently, rotation-invariant structural descriptors are learned from global PPF signatures and leveraged to incorporate awareness of global 3D structures into local descriptors. A Transformer~\cite{vaswani2017attention} architecture is further added, yielding a Vision Transformer (ViT)~\cite{dosovitskiy2020image} architecture to incorporate global awareness of geometric context. Finally, dense point descriptors are obtained by interpolation, and the coarse-to-fine mechanism proposed in CoFiNet~\cite{yu2021cofinet} is extended to extract reliable correspondences from our rotation-invariant and globally-aware descriptors for point cloud registration.


To the best of our knowledge, \OURS{} is the first to learn both rotation-invariant and globally-aware descriptors for point cloud registration. Our contributions are summarized as:
\begin{itemize}

\item We propose an end-to-end pipeline that guarantees the rotational invariance of globally-aware descriptors by design and extracts coarse-to-fine correspondences for point cloud registration.

\item We propose global PPF signatures to provide a node-specific description of the entire scene in a rotation-invariant fashion and further learn global structural descriptors from them to incorporate global structural awareness into local descriptors.

\item We empirically show the effectiveness of rotational invariance and global awareness on both object- and scene-level data.

\end{itemize}

%% file: sections/2related_work.tex
\section{Related Work}
\label{Sec.2}
\subsection{Rotation-Invariant Descriptors}
\subsubsection{Handcrafted Rotation-Invariant Descriptors}
Handcrafted rotation-invariant descriptors~\cite{rusu2008aligning,rusu2009fast,drost2010model,tombari2010unique,guo2013rotational} have been widely explored in 3D by researchers before the popularity of deep neural networks. To guarantee the invariance under rotations, many handcrafted local descriptors~\cite{tombari2010unique,guo2013rotational} rely on an estimated local reference frame~(LRF), which is typically based on the covariance analysis of the local surface, to transform local patches to a defined canonical representation. The major drawback of LRF is its non-uniqueness. The constructed rotational invariance is therefore fragile and sensitive to noise. As a result, the attention shifts to those LRF-free approaches~\cite{rusu2008aligning,rusu2009fast,drost2010model}. These methods focus on mining the rotation-invariant components of local surfaces and using them to represent the local geometry. Given a point of interest and its adjacent points within the vicinity area, PPF~\cite{drost2010model} describes each pairwise relationship using Euclidean distances and angles among point vectors and normals. In a similar way, PFH~\cite{rusu2008aligning} and FPFH~\cite{rusu2009fast} encode the geometry of the local surface using the histogram of pairwise geometrical properties. Although these handcrafted descriptors are rotation-invariant by design, all of them are far from satisfactory to be applied in real scenarios with complicated geometry and severe noise. 

\begin{figure*}[t!]
	\centering
	\includegraphics[width=0.95\linewidth]{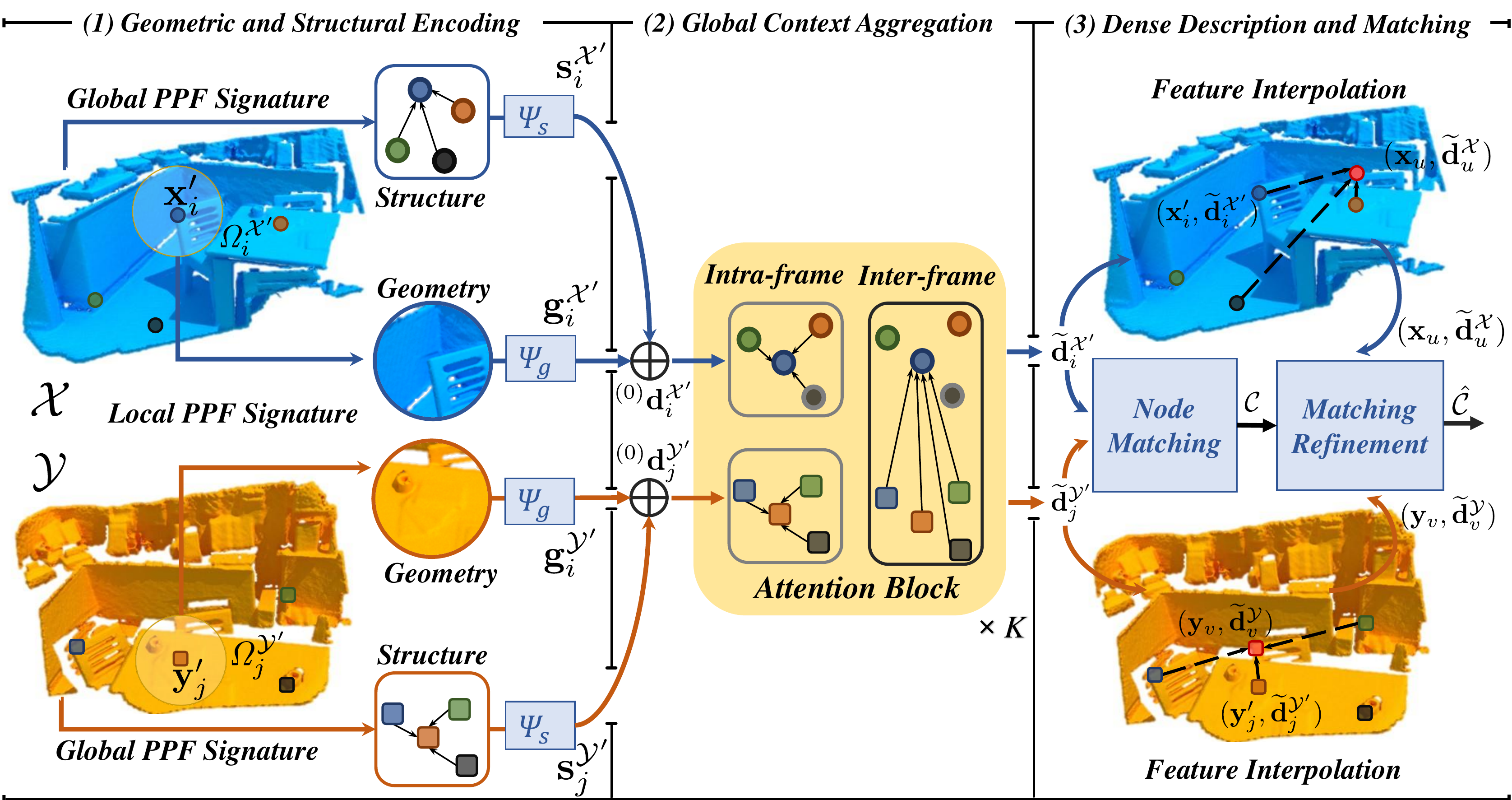}
	\caption{\textbf{Method Overview.} Point cloud $\mathcal{X}$ and $\mathcal{Y}$ are processed in the same way, and we only explain for $\mathcal{X}$ hereafter. \textbf{(1)}~Local and global PPF signatures are computed for each node $\mathbf{x}_i$, which is sparsely sampled from $\mathcal{X}$. Local geometry and global structures are encoded into descriptors $\mathbf{g}^{\mathcal{X}^\prime}_i$ and $\mathbf{s}^{\mathcal{X}^\prime}_i$ by PointNet~\cite{qi2017pointnet} $\Psi_g$ and $\Psi_s$, respectively. \textbf{(2)}~$\mathbf{s}^{\mathcal{X}^\prime}_i$ joins $\mathbf{g}^{\mathcal{X}^\prime}_i$ with global 3D structures via element-wise addition, yielding a globally-informed descriptor $^{\scriptsize{(0)}}\mathbf{d}^{\mathcal{X}^\prime}_i$. A stack of $K$ attention blocks is leveraged, where intra- and inter-frame geometric context is globally incorporated, resulting in a globally-aware descriptor $\widetilde{\mathbf{d}}^{\mathcal{X}^\prime}_i$. \textbf{(3)}~Descriptor $\widetilde{\mathbf{d}}^{\mathcal{X}}_u$ of every point $\mathbf{x}_u\in \mathcal{X}$ is obtained via interpolation. Node correspondence set $\mathcal{C}$ is retrieved in the Node Matching Module~(Fig.~\ref{fig:c2f}(a)). In the Matching Refinement Module~(Fig.~\ref{fig:c2f}(b)), point correspondence set $\hat{\mathcal{C}}$ is extracted according to $\mathcal{C}$ and point descriptors. All the descriptors are invariant to rotations by design.}
	\label{fig:overview}

\end{figure*}

\subsubsection{Learning-based Rotation-Invariant Descriptors}

 Recently, many deep learning-based methods~\cite{deng2018ppf,gojcic2019perfect,ao2021spinnet} make the attempt to learn descriptors in a rotation-invariant fashion. As a pioneer, PPF-FoldNet~\cite{deng2018ppf} encodes PPF patches into embeddings, from which a FoldingNet~\cite{yang2018foldingnet} decoder reconstructs the input. Correspondences are extracted from the rotation-invariant embeddings for registration. Different from PPF-FoldNet~\cite{deng2018ppf} that learns from handcrafted LRF-free descriptors, 3DSN~\cite{gojcic2019perfect} leverages LRF, which transforms local patches around interest points to defined canonical representations, to enhance the robustness of learned descriptors against rotations. Similarly, SpinNet~\cite{ao2021spinnet} and Graphite~\cite{saleh2020graphite,saleh2022bending} align local patches according to the defined axes before learning descriptors from them. However, all those methods are limited by their locality, i.e., their descriptors are only learned from the local region where their rotational invariance is defined. Those descriptors are blind to the global context and are therefore less distinctive. Without relying on rotation-invariant handcrafted features, YOHO~\cite{wang2021you} leverages an icosahedral group to learn a group of rotation-equivariant descriptors for each point. Rotating the input point cloud will permute the descriptors within the group, and rotational invariance is achieved by max-pooling over the group. However, its rotational equivariance is fragile in practice, as the finite rotation group cannot span the infinite rotation space. Additionally, expanding a single descriptor to a group damages efficiency. In object-centric registration, recent methods~\cite{yew2020rpm,pan2021robust,drost2010model} strengthen the rotational invariance in their learned descriptors by concatenating rotation-invariant descriptors, e.g., PPF~\cite{drost2010model}, with their rotation-variant input. However, as shown in Tab.~\ref{tab:object}, the registration performance of those methods still drops severely when facing large rotations~\cite{pan2021robust}.

\subsection{Globally-Aware Descriptors}
PPF, as an example, has been made semi-global before the widespread of deep neural networks for different tasks~\cite{birdal2015point,birdal2017cad,drost2010model,hinterstoisser2016going}.  With the widespread of deep neural networks, PPFNet~\cite{deng2018ppf} makes the first attempt to incorporate learned global context into their learned descriptors. However, their descriptors are rotation-variant in nature, as the absolute coordinates and PPF features are concatenated as input. Moreover, naively leveraging a max-pooling operator for global awareness largely neglects global information beyond each local patch. Predator~\cite{huang2021predator} leverages attention~\cite{vaswani2017attention} mechanism in a point cloud registration method to strengthen their descriptors with learned global context. Global information is incorporated from the same and the opposite frame, by interleaving Edge Conv-based~\cite{wang2019dynamic} self-attention modules and  Transformer-based~\cite{vaswani2017attention} cross-attention modules, respectively. Similarly, Yu et al.~\cite{yu2021cofinet} interleave Transformer-based~\cite{vaswani2017attention} self- and cross-attention modules for learning globally-aware descriptors. Such a paradigm is also leveraged in the most recent works~\cite{li2022lepard,yew2022regtr,qin2022geometric} for incorporating global awareness into local descriptors. However, these methods ignore the inherent rotational invariance of their learned descriptors. As a result, rotational invariance is learned through data augmentation during training, which is intricate for large rotations and adds significant capacity requirements to the deep model.

%% file: sections/3method.tex
\section{Method}
\label{Sec.3}
\subsection{Problem Statement}
\label{Sec.3_1}
We aim at recovering the rigid transformation $\mathbf{T} = \{\mathbf{R}\in \mathcal{SO}(3), \mathbf{t}\in \mathbb{R}^{3}\}$ that best aligns two partially-overlapping point clouds $\mathcal{X}=\{\mathbf{x}_{1}, \dots, \mathbf{x}_{N}\}$ and $\mathcal{Y} = \{\mathbf{y}_{1}, \dots, \mathbf{y}_{M}\}$.
We follow the paradigm of those correspondence-based models~\cite{deng2018ppfnet,choy2019fully,bai2020d3feat,ao2021spinnet,huang2021predator,yu2021cofinet,qin2022geometric,yew2022regtr,li2022lepard}, where transformation is solved based on a putative correspondence set $\hat{\mathcal{C}}$ according to:
\begin{equation}
\arg\min_{\mathbf{R}, \mathbf{t}}\sum_{(\mathbf{x}_i, \mathbf{y}_j)\in \hat{\mathcal{C}}} \lVert \mathbf{R}\cdot\mathbf{x}_i + \mathbf{t} - \mathbf{y}_j\rVert_2^2,
\label{eq:solve}
\end{equation}
where $\lVert\cdot \rVert_{2}$ represents the Euclidean norm, and the correspondence set $\hat{\mathcal{C}}$ is established by matching points according to their associated descriptors. In this paper, we focus on learning more powerful descriptors that are inherently rotation-invariant and globally aware. By combining the coarse-to-fine matching mechanism~\cite{yu2021cofinet}, our descriptors lead to more reliable correspondences and thus better registration performance. An overview of the \OURS{} pipeline can be found in Fig.~\ref{fig:overview}.

\subsection{Learning Rotation-Invariant Descriptors from Local Geometry}
\label{Sec.3_2}
\begin{figure}[t!]
	\centering
	\includegraphics[width=1.0\linewidth]{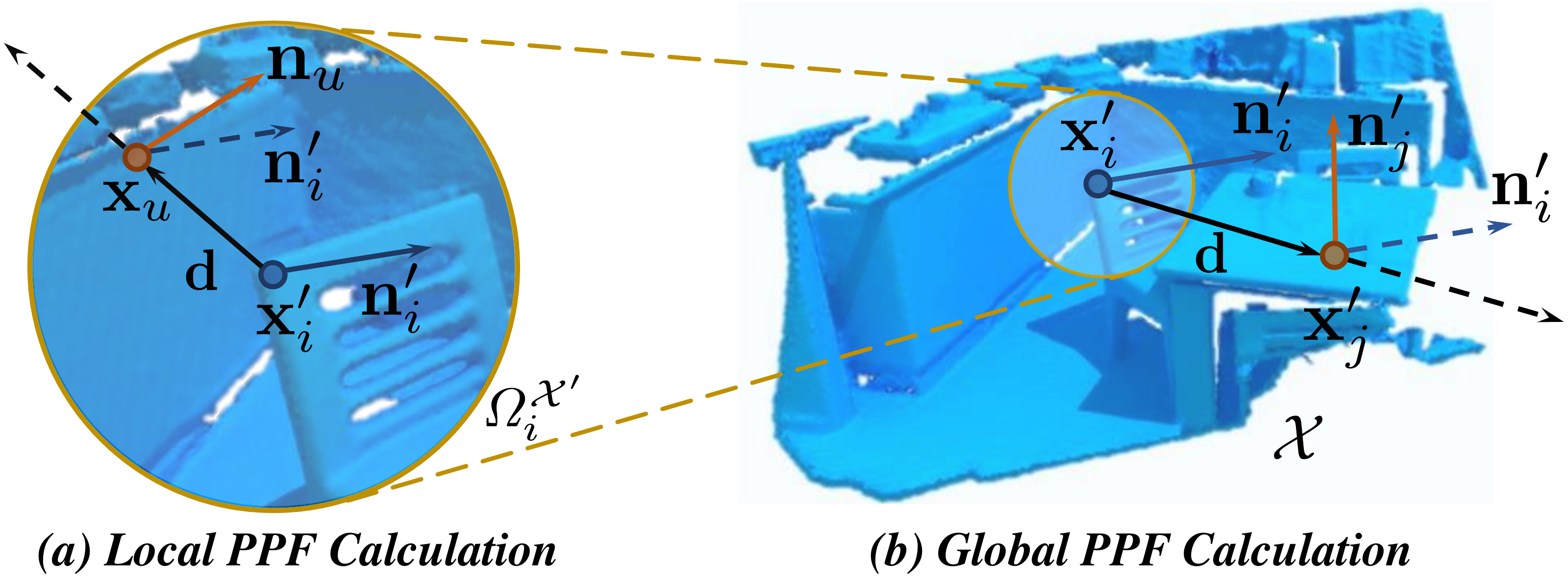}
	\caption{\textbf{Illustration of PPF Calculation.} $\mathbf{n}$ and $\mathbf{n}^\prime$ denote normals. (a) shows the local PPF of a point $\mathbf{x}_u\in \Omega^{\mathcal{X}^\prime}_i$ with respect to node $\mathbf{x}^\prime_i$. (b) shows the global PPF setup of a node $\mathbf{x}^\prime_j$ sampled from $\mathcal{X}$ with respect to node $\mathbf{x}^\prime_i$. 
	}
	\label{fig:ppf}

\end{figure}

The first step of our method is the rotation-invariant encoding of geometry within local areas.  In the following, we will explain it on the example of $\mathcal{X}$. Encoding is done in exactly the same way for $\mathcal{Y}$. Firstly, $N^{\prime}$ nodes $\mathcal{X}^{\prime}=\{\mathbf{x}^{\prime}_{1},\mathbf{x}^{\prime}_{2},\cdots,\mathbf{x}^{\prime}_{N^{\prime}}\}$ are sampled out of $N$ points via Farthest Point Sampling~\cite{qi2017pointnet++}. For each node $\mathbf{x}^{\prime}_i \in \mathcal{X}^{\prime}$, its support area $\Omega^{\mathcal{X}^\prime}_i$ can be defined by a radius $r \in \mathbb{R}$, which is demonstrated as: 
\begin{equation}
\Omega^{\mathcal{X}^\prime}_i = \{\mathbf{x}_u \in \mathcal{X}\big| {\lVert \mathbf{x}^{\prime}_i - \mathbf{x}_u\rVert}_2 < r\}.
\label{eq:support}
\end{equation}
 Each support area is represented with a set of rotation-invariant PPFs~\cite{drost2010model}.  As shown in Fig.~\ref{fig:ppf}(a), for node $\mathbf{x}^{\prime}_i$, normal $\mathbf{n}^\prime_i$ of $\mathbf{x}^{\prime}_i$ and $\mathbf{n}_u$ of each point $\mathbf{x}_u\in \Omega^{\mathcal{X}^\prime}_i$ are estimated~\cite{hoppe1992surface}, and the local PPF signature of $\mathbf{x}^\prime_i$ is represented as a set of PPFs:

\begin{equation}
\mathcal{S}_{l}(\mathbf{x}^{\prime}_i| \Omega^{\mathcal{X}^\prime}_i) = \{\mathbf{\xi}(\mathbf{x}_u, \mathbf{n}_u |\mathbf{x}^{\prime}_i, \mathbf{n}^{\prime}_i)\big|  \mathbf{x}_u \in \Omega^{\mathcal{X}^\prime}_i\},
\label{eq:local_signature}
\end{equation}
\noindent with each PPF defined as:

\begin{equation}
\mathbf{\xi}(\mathbf{x}_u, \mathbf{n}_u |\mathbf{x}^{\prime}_i, \mathbf{n}^{\prime}_i) = (\lVert \mathbf{d} \rVert_2, \angle(\mathbf{n}^{\prime}_i, \mathbf{d}), \angle(\mathbf{n}_u, \mathbf{d}), \angle(\mathbf{n}^{\prime}_i, \mathbf{n}_u)),
\label{eq:ppf}
\end{equation}
where $\mathbf{d}$ represents the vector between $\mathbf{x}^{\prime}_i$ and $\mathbf{x}_u$, and $\angle$ computes the angle between two vectors $\mathbf{v}_1$ and $\mathbf{v}_2$, following the way in~\cite{birdal2015point,deng2018ppfnet}:

\begin{equation}
\angle(\mathbf{v}_1, \mathbf{v}_2) = atan2( \lVert \mathbf{v}_1 \times \mathbf{v}_2 \rVert_{2}, \mathbf{v}_1 \cdot \mathbf{v}_2).
\label{eq:atan2}
\end{equation}

Then, we leverage PointNet~\cite{qi2017pointnet} to project each local PPF signature to a \textit{c}-dimension local geometric descriptor:

\begin{equation}
\mathbf{g}^{\mathcal{X}^\prime}_i = \Psi_g(\mathcal{S}_{l}(\mathbf{x}^{\prime}_i | \Omega^{\mathcal{X}^\prime}_i))\in \mathbb{R}^{c}, \qquad 1 \leq i \leq N^{\prime},
\label{eq:local_descriptor}
\end{equation}

\noindent where $\Psi_g$ stands for a PointNet~\cite{qi2017pointnet} model shared across all the support areas, and \textit{c} is the dimension of learned local descriptors. As a result, each support area is described by a rotation-invariant geometric descriptor of length \textit{c}.

\subsection{Learning Rotation-Invariant Descriptors from Global 3D Structures}
\label{Sec.3_3}
 The learned geometric descriptor $\mathbf{g}^{\mathcal{X}^\prime}_i$, defined in Eq.~\ref{eq:local_descriptor}, is conditioned only on its support area $\Omega^{\mathcal{X}^\prime}_i$. Consequently, it lacks awareness of the global context and is less distinctive for correspondence search. We consider this the main reason why existing rotation-invariant methods~\cite{deng2018ppfnet,gojcic2019perfect,ao2021spinnet,wang2021you} fail to compete with rotation-variant but globally-aware approaches~\cite{huang2021predator,yu2021cofinet,li2022lepard}. To address this issue, we propose to enrich local descriptors with global structural cues learned from our global PPF signatures that are invariant to rotations by design.

The design of global PPF signatures is inspired by the handcrafted PPF which is widely used for describing local geometry. For each node $\mathbf{x}^{\prime}_i$ with normal $\mathbf{n}^{\prime}_i$~$(1\leq i \leq N^{\prime})$, we compute the structural relationship of every other node $\mathbf{x}^{\prime}_j\in \mathcal{X}^\prime$ w.r.t. it (see Fig.~\ref{fig:ppf}(b)) by:
\begin{equation}
\mathcal{S}_g(\mathbf{x}^{\prime}_i | \mathcal{X}^{\prime}) = \{ \mathbf{\xi}(\mathbf{x}^{\prime}_j, \mathbf{n}^{\prime}_j |\mathbf{x}^{\prime}_i,  \mathbf{n}^{\prime}_i)\big| \mathbf{x}^\prime_j \in \mathcal{X}^\prime, j\neq i\},
\label{eq:condition}
\end{equation}

\noindent which we define as the global PPF signature of node $\mathbf{x}^{\prime}_i$. Similar to the conventional PPF, the obtained global PPF signatures are rotation-invariant by design. However, the global PPF signatures are unordered as well. Besides, as the global PPF signatures are conditioned on the whole scene represented by sparse nodes, they can be sensitive to partial overlap, i.e., although some nodes can be occluded in $\mathcal{Y}$, they still contribute to the structural awareness of $\mathbf{x}^{\prime}_i$. 
Therefore, we further leverage a second PointNet~\cite{qi2017pointnet} architecture $\Psi_s$ to address both issues simultaneously. The network $\Psi_s$ projects each global PPF signature to a \textit{c}-dimension structural descriptor. This successfully eliminates the inherent unordered property of the global PPF signatures and provides more robustness against partial overlap in real scenes. We denote the obtained structural descriptors as:
 
\begin{equation}
\mathbf{s}^{\mathcal{X}^\prime}_i = \Psi_s(\mathcal{S}_g(\mathbf{x}^{\prime}_i |\mathcal{X}^{\prime}))\in \mathbb{R}^{c}, \qquad 1 \leq i \leq N^{\prime}.
\label{eq:rige}
\end{equation}

\noindent Each global structural descriptor $\mathbf{s}^{\mathcal{X}^\prime}_i$ will be used to inform its corresponding local geometric descriptor $\mathbf{g}^{\mathcal{X}^\prime}_i$ with global structural information from 3D space. 

\subsection{Rotation-Invariant Global Awareness}
\label{Sec.3_4}
\subsubsection{Incorporating Global Information from 3D Structures}
Following the examples of \cite{sarlin2020superglue,huang2021predator,yu2021cofinet}, we interleave self- and cross-attention for intra- and inter-frame global context, respectively. However, the standard attention~\cite{vaswani2017attention} lacks the awareness of global 3D structures, as it is based purely on the similarity of learned geometry. To this end, we inform each learned local geometric descriptor $\mathbf{g}^{\mathcal{X}^\prime}_i$~($1\leq i \leq N^{\prime}$) and $\mathbf{g}^{\mathcal{Y}^\prime}_j$~($1\leq j \leq M^{\prime}$) with global structural cues encoded in corresponding global structural descriptor $\mathbf{s}^{\mathcal{X}^\prime}_i$ and $\mathbf{s}^{\mathcal{Y}^\prime}_j$, respectively. The obtained globally-informed descriptors are calculated as
$^{\scriptsize{(0)}}\mathbf{d}^{\mathcal{X}^\prime}_i = \mathbf{g}^{\mathcal{X}^\prime}_i \oplus \mathbf{s}^{\mathcal{X}^\prime}_i$ and $^{\scriptsize{(0)}}\mathbf{d}^{\mathcal{Y}^\prime}_j = \mathbf{g}^{\mathcal{Y}^\prime}_j \oplus   \mathbf{s}^{\mathcal{Y}^\prime}_j$, where $\oplus$ is the element-wise addition. 

\begin{figure*}[h]
	\centering
	\includegraphics[width=0.75\linewidth]{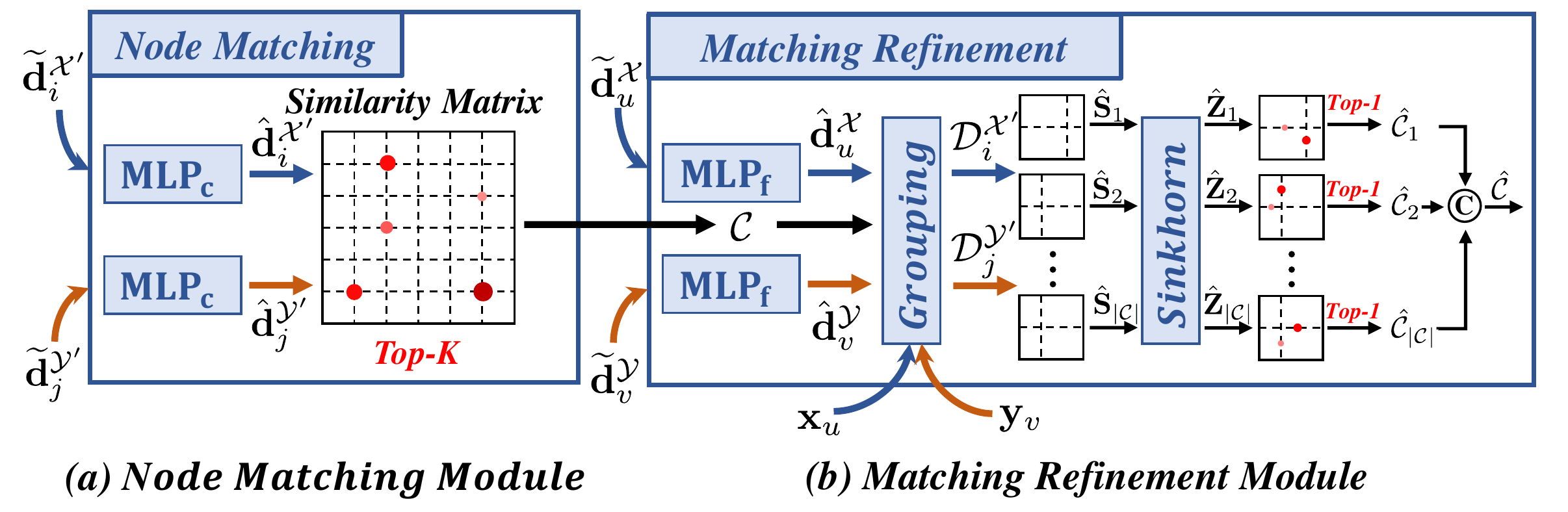}
	\caption{\textbf{Illustration of Coarse-to-Fine Correspondence Extraction.} In (a), nodes from two frames are matched according to the similarity of the MLP-projected descriptors, and node correspondences with Top-\textit{K} highest scores are selected. In (b), according to  Eq.~\ref{eq:vicinity}, each node is assigned with a group of neighbor points, together with their associated MLP-projected descriptors. For each node correspondence, the similarity between their neighbor points is computed. The resulting similarity matrix is normalized by Sinkhorn~\cite{sinkhorn1967concerning} algorithm. A point correspondence set is extracted from each normalized matrix, and the final point correspondence set is constructed as the union of all the individual ones. 
	}
	\label{fig:c2f}
\end{figure*}

\subsubsection{Global Intra-Frame Aggregation of Geometric Context}

A stack of \textit{K} attention blocks operates on globally-informed descriptors to exchange learned geometric information among nodes. Each attention block has an intra-frame module followed by an inter-frame module.
 
Taking node $\mathbf{x}^\prime_i\in \mathcal{X}^\prime$ as an example, we detail the computation of the intra-frame module inside the $l^{th}$~($1\leq l \leq K$) attention block hereafter. Learnable matrices ${^{\scriptsize{(l)}}}\mathbf{W}_q$, ${^{\scriptsize{(l)}}}\mathbf{W}_k$, and ${^{\scriptsize{(l)}}}\mathbf{W}_v \in \mathbb{R}^{c\times c}$ are introduced to linearly project ${^{\scriptsize{(l-1)}}}\mathbf{d}^{\mathcal{X}^\prime}_i$ to \textit{query}, \textit{key}, and \textit{value} with:

\begin{equation}
\begin{aligned}
^{(l)}\mathbf{q}^{\mathcal{X}^\prime}_i = {^{\scriptsize{(l)}}}\mathbf{W}_q \cdot {^{\scriptsize{(l-1)}}}{\mathbf{d}^{\mathcal{X}^\prime}_i},
\\ ^{(l)}\mathbf{k}^{\mathcal{X}^\prime}_i = {^{\scriptsize{(l)}}}\mathbf{W}_k \cdot {^{\scriptsize{(l-1)}}}{\mathbf{d}^{\mathcal{X}^\prime}_i},
\\ ^{(l)}\mathbf{v}^{\mathcal{X}^\prime}_i = {^{\scriptsize{(l)}}}\mathbf{W}_v \cdot {^{\scriptsize{(l-1)}}}{\mathbf{d}^{\mathcal{X}^\prime}_i},
\end{aligned}
\label{eq:project}
\end{equation}

\noindent respectively, where ${^{(l)}\mathbf{q}_i^{\mathcal{X}^\prime}}$ and ${^{(l)}\mathbf{k}_i^{\mathcal{X}^\prime}}$ are used for retrieving similar nodes, and ${^{(l)}\mathbf{v}_i^{\mathcal{X}^\prime}}$ encodes the context for aggregation.

The attention~\cite{vaswani2017attention} is defined on a node set $\mathcal{S}\in 
\{\mathcal{X}^\prime, \mathcal{Y}^\prime\}$:
\begin{equation}
{^{(l)}\mathbf{a}^{{\mathcal{X}^\prime}\leftarrow \mathcal{S}}_i} = \text{softmax}([ {^{(l)}a_{i}^{1}}, {^{(l)}a_{i}^{2}},\cdots,{^{(l)}a_{i}^{|\mathcal{S}|}}])^{T}/\sqrt{c} \in \mathbb{R}^{|\mathcal{S}|},\\ 
\label{eq:attention}
\end{equation}
\noindent where $^{(l)}a_{i}^{j}$ is calculated as $^{(l)}a_{i}^{j} = (^{(l)}\mathbf{q}^{\mathcal{X}^\prime}_{i})^{T} \cdot  {^{(l)}\mathbf{k}^{\mathcal{S}}_{j}}$~($1\leq j \leq |\mathcal{S}|$), and $|\cdot|$ denotes the set cardinality. The message $^{(l)}\mathbf{m}^{\mathcal{X}^\prime \leftarrow \mathcal{S}}_i\in \mathbb{R}^c$, which flows from set $\mathcal{S}$ to node $\mathbf{x}^\prime_i\in \mathcal{X}^\prime$, is calculated as:
\vspace{-0.1cm}
\begin{equation}
^{(l)}\mathbf{m}^{\mathcal{X}^\prime \leftarrow \mathcal{S}}_i = [^{(l)}\mathbf{v}^{\mathcal{S}}_1,^{(l)}\mathbf{v}^{\mathcal{S}}_2,\cdots, ^{(l)}\mathbf{v}^{\mathcal{S}}_{|\mathcal{S}|}] \cdot ^{(l)}\mathbf{a}^{{\mathcal{X}^\prime}\leftarrow \mathcal{S}}_i \in \mathbb{R}^c.\\ 
\label{eq:message}
\end{equation}

We globally aggregate the intra-frame learned geometry with:
\begin{equation}
{^{\scriptsize{(l)}}\overline{\mathbf{d}}^{\mathcal{X}^\prime}_i} = {^{\scriptsize{(l-1)}}\mathbf{d}^{\mathcal{X}^\prime}_i} + \text{MLP}(\big[{^{\scriptsize{(l-1)}}\mathbf{d}^{\mathcal{X}^\prime}_i} ,\mathbf{m}^{\mathcal{X}^\prime \leftarrow \mathcal{S}}_i\big]),\\
\\
\label{eq:aggregation}
\end{equation}

\noindent where $\text{MLP}$ is a multilayer perceptron with $\mathcal{S} = \mathcal{X}^\prime$. For node  $\mathbf{y}^\prime_j\in \mathcal{Y}^\prime$, ${^{\scriptsize{(l)}}\overline{\mathbf{d}}^{\mathcal{Y}^\prime}_j}$ is calculated in the same way according to Eq.~\ref{eq:aggregation}, but with $\mathcal{S} = \mathcal{Y}^\prime$.

\subsubsection{Global Inter-Frame Fusion of Geometric Context}
For the $l^{th}~(1\leq l \leq K)$ attention block, the inter-frame module takes as input the output of the intra-frame module, i.e., ${^{\scriptsize{(l)}}\overline{\mathbf{d}}^{\mathcal{X}^\prime}_i}$ and ${^{\scriptsize{(l)}}\overline{\mathbf{d}}^{\mathcal{Y}^\prime}_j}$. Taking node $\mathbf{x}^\prime_i\in \mathcal{X}^\prime$ as an example, similar to Eq.~\ref{eq:project}, ${^{\scriptsize{(l)}}}\overline{\mathbf{d}}^{\mathcal{X}^\prime}_i$ is linearly projected by learnable matrices ${^{\scriptsize{(l)}}}\overline{\mathbf{W}}_q$, ${^{\scriptsize{(l)}}}\overline{\mathbf{W}}_k$, and ${^{\scriptsize{(l)}}}\overline{\mathbf{W}}_v \in \mathbb{R}^{c\times c}$ :

\begin{equation}
\begin{aligned}
^{(l)}\overline{\mathbf{q}}^{\mathcal{X}^\prime}_i = {^{\scriptsize{(l)}}}\overline{\mathbf{W}}_q \cdot {^{\scriptsize{(l)}}}{\overline{\mathbf{d}}^{\mathcal{X}^\prime}_i},
\\ ^{(l)}\overline{\mathbf{k}}^{\mathcal{X}^\prime}_i = {^{\scriptsize{(l)}}}\overline{\mathbf{W}}_k \cdot {^{\scriptsize{(l)}}}{\overline{\mathbf{d}}^{\mathcal{X}^\prime}_i},
\\ ^{(l)}\overline{\mathbf{v}}^{\mathcal{X}^\prime}_i = {^{\scriptsize{(l)}}}\overline{\mathbf{W}}_v \cdot {^{\scriptsize{(l)}}}{\overline{\mathbf{d}}^{\mathcal{X}^\prime}_i},
\end{aligned}
\label{eq:project2}
\end{equation}

\noindent upon which ${^{(l)}\overline{\mathbf{a}}^{{\mathcal{X}^\prime}\leftarrow \mathcal{S}}_i}$ and $^{(l)}\overline{\mathbf{m}}^{\mathcal{X}^\prime \leftarrow \mathcal{S}}_i$ are computed following Eq.~\ref{eq:attention} and Eq.~\ref{eq:message}, respectively,  with $\mathcal{S}=\mathcal{Y}^\prime$. Finally, the geometric context from the opposite frame, i.e., the node set $\mathcal{Y}^\prime$, is fused to node $\mathbf{x}^\prime_i$:

\begin{equation}
{^{\scriptsize{(l)}}\mathbf{d}^{\mathcal{X}^\prime}_i} = {^{\scriptsize{(l)}}\overline{\mathbf{d}}^{\mathcal{X}^\prime}_i} + \text{MLP}(\big[{^{\scriptsize{(l)}}\overline{\mathbf{d}}^{\mathcal{X}^\prime}_i} ,\overline{\mathbf{m}}^{\mathcal{X}^\prime \leftarrow \mathcal{S}}_i\big]),\\
\\
\label{eq:aggregation2}
\end{equation}
with $\mathcal{S} = \mathcal{Y}^\prime$. For node  $\mathbf{y}^\prime_j\in \mathcal{Y}^\prime$, ${^{\scriptsize{(l)}}\mathbf{d}^{\mathcal{Y}^\prime}_j}$ is calculated in the same way according to Eq.~\ref{eq:aggregation2}, but with $\mathcal{S} = \mathcal{X}^\prime$.

Since all the operation is performed in feature space, the rotation-invariance of $^{\scriptsize{(0)}}\mathbf{d}^{\mathcal{X}^\prime}_i$ remains in all $^{\scriptsize{(l)}}\overline{\mathbf{d}}^{\mathcal{X}^\prime}_i$ and $^{\scriptsize{(l)}}\mathbf{d}^{\mathcal{X}^\prime}_i$ with $1 \leq l \leq K$. As a result, the obtained globally-aware descriptor $\widetilde{\mathbf{d}}^{\mathcal{X}^\prime}_i \defeq {^{\scriptsize{(K)}}\mathbf{d}^{\mathcal{X}^\prime}_i}$ is rotation-invariant by design. Similarly, globally-aware descriptor $\widetilde{\mathbf{d}}^{\mathcal{Y}^\prime}_j \defeq {^{\scriptsize{(K)}}\mathbf{d}^{\mathcal{Y}^\prime}_j}$ is also rotation-invariant for each $\mathbf{y}^\prime_j\in \mathcal{Y}$.

\subsection{Rotation-Invariant Dense Description}
\label{Sec.3_5}
Until here, we have successfully incorporated global awareness into learned local descriptors of nodes without sacrificing the inherent rotational invariance. The aforementioned repeatability issue of sparsely sampled nodes, however, still remains. To address this issue, we leverage the coarse-to-fine strategy proposed in~\cite{yu2021cofinet}, where nodes are first matched according to the overlap ratios of their vicinities, and point correspondences are then extracted from the vicinities of matched nodes. As the first step, dense point descriptors are generated via interpolation. For each point $\mathbf{x}_u\in \mathcal{X}$, we find its \textit{k}-nearest neighbor nodes in $\mathcal{X}^\prime$ according to their Euclidean distance. The descriptor $\widetilde{\mathbf{d}}^{\mathcal{X}}_u$ of point $\mathbf{x}_u$ can be interpolated as:

\begin{equation}
\widetilde{\mathbf{d}}^{\mathcal{X}}_u = \sum_{i=1}^{k} w^u_i\cdot \widetilde{\mathbf{d}}^{\mathcal{X}^\prime}_i, \qquad \text{with} \quad w^u_i = \frac{1/d^u_i}{\sum_{l=1}^{k}1/d^u_l},
\label{eq:interpolation}
\end{equation}
where $d^u_l$ depicts the Euclidean distance of point $\mathbf{x}_u$ to its $\mathit{l}^{th}$ nearest node in geometry space. Point descriptor $\widetilde{\mathbf{d}}^{\mathcal{Y}}_v$ of $\mathbf{y}_v\in \mathcal{Y}$ is calculated in the same way. As the interpolation coefficients are only related to Euclidean distance, the obtained point descriptors remain invariant to rotations.

\subsection{Coarse-to-Fine Correspondence Extraction}
\label{Sec.3_6}
The coarse-to-fine mechanism~\cite{yu2021cofinet} is leveraged to extract correspondences from our obtained node and point descriptors. We first project $\widetilde{\mathbf{d}}^{\mathcal{X}^\prime}_i$ and $\widetilde{\mathbf{d}}^{\mathcal{X}}_u$ by using two individual multilayer perceptrons~($\text{MLP}$), which provides $\hat{\mathbf{d}}^{\mathcal{X}^\prime}_i$ and $\hat{\mathbf{d}}^{\mathcal{X}}_u$ in Fig.~\ref{fig:c2f}(a) and (b), respectively. We also project descriptors from point cloud $\mathcal{Y}$ to $\hat{\mathbf{d}}^{\mathcal{Y}^\prime}_j$ and $\hat{\mathbf{d}}^{\mathcal{Y}}_v$. On the coarse level, as shown in Fig.~\ref{fig:c2f}(a), the similarity between node $\mathbf{x}^\prime_i\in \mathcal{X}^\prime$ and $\mathbf{y}^\prime_j\in \mathcal{Y}^\prime$ is calculated as $1 / \lVert \hat{\mathbf{d}}^{\mathcal{X}^\prime}_i$ - $\hat{\mathbf{d}}^{\mathcal{Y}^\prime}_j \rVert_2$. As the following step, Top-\textit{K} node correspondences with the highest similarity values are sampled, resulting in the node correspondence set $\mathcal{C}$ with $|\mathcal{C}|$ correspondences. In "Grouping" of Fig.~\ref{fig:c2f}(b), vicinities $(\mathcal{V}^{\mathcal{X}^\prime}_i, \mathcal{V}^{\mathcal{Y}^\prime}_j)$ of coarse correspondence $C_l \defeq (\mathbf{x}^{\prime}_i, \mathbf{y}^{\prime}_j) \in \mathcal{C}$ are collected by the point-to-node assignment~\cite{li2019usip,yu2021cofinet}, i.e., assigning points to their nearest nodes in geometry space. For node $\mathbf{x}^{\prime}_i$, its vicinity $\mathcal{V}^{\mathcal{X}^\prime}_i$ and the associated descriptor group $\mathcal{D}^{\mathcal{X}^\prime}_i$ can be defined as:
\begin{equation}
\left\{
\begin{array}{l}
     \mathcal{V}^{\mathcal{X}^\prime}_i = \{\mathbf{x}_u \in \mathcal{X} \big| \lVert\mathbf{x}_u - \mathbf{x}^{\prime}_i\rVert_2 < \lVert \mathbf{x}_u - \mathbf{x}^{\prime}_j\rVert_2, \forall j \neq i \},\\
     \\
     \mathcal{D}^{\mathcal{X}^\prime}_i=\{\hat{\mathbf{d}}^{\mathcal{X}}_u\big| \hat{\mathbf{d}}^{\mathcal{X}}_u \leftrightarrow \mathbf{x}_u \ with\  \mathbf{x}_u \in \mathcal{V}^{\mathcal{X}^\prime}_i\},
\end{array}
\right.
\label{eq:vicinity}
\end{equation}
where $\hat{\mathbf{d}}^{\mathcal{X}}_u \leftrightarrow \mathbf{x}_u$ denotes that $\hat{\mathbf{d}}^{\mathcal{X}}_u$ is the descriptor associated to point $\mathbf{x}_u$. $\mathcal{V}^{\mathcal{Y}^\prime}_j$ and $\mathcal{D}^{\mathcal{Y}^\prime}_j$ are defined in the same way for nodes $\mathbf{y}^\prime_j\in \mathcal{Y}^\prime$. Finally, we present the similarity of $(\mathcal{D}^{\mathcal{X}^\prime}_i, \mathcal{D}^{\mathcal{Y}^\prime}_j)$ as a matrix $\hat{\mathbf{S}}_l\in \mathbb{R}^{|\mathcal{D}^{\mathcal{X}^\prime}_i|\times|\mathcal{D}^{\mathcal{Y}^\prime}_j|}$, where each entry is calculated as $\hat{\mathbf{S}}_l^{u, v} = (\hat{\mathbf{d}}^{\mathcal{X}}_u)^T \cdot \hat{\mathbf{d}}^\mathcal{Y}_v$, with $\hat{\mathbf{d}}^{\mathcal{X}}_u\in \mathcal{D}^{\mathcal{X}^\prime}_i$ and $\hat{\mathbf{d}}^{\mathcal{Y}}_v\in \mathcal{D}^{\mathcal{Y}^\prime}_j$. To deal with partial overlap, we follow the slack idea~\cite{sarlin2020superglue} and augment $\hat{\mathbf{S}}_l$ with an additional row and an additional column filled with the same learnable parameter $\alpha$. In "Sinkhorn" of Fig.~\ref{fig:c2f}(b), each augmented similarity matrix is normalized to a confidence matrix $\hat{\mathbf{Z}}_l\in \mathbb{R}^{|\mathcal{D}^{\mathcal{X}^\prime}_i + 1|\times|\mathcal{D}^{\mathcal{Y}^\prime}_j + 1|}$, which is a non-negative matrix with every row and every column summing to 1, with the Sinkhorn~\cite{sinkhorn1967concerning} algorithm. From $\hat{\mathbf{Z}}_l$ we extract the point correspondence set $\hat{\mathcal{C}}_l$ as the maximum confidence individually for each row and column. The union of all $\hat{\mathcal{C}}_l$~($1\leq l \leq |\mathcal{C}|$) constructs the final point correspondence set $\hat{\mathcal{C}}$, which we use for registration.

\begin{figure*}[h]
	\centering
	\includegraphics[width=0.9\linewidth]{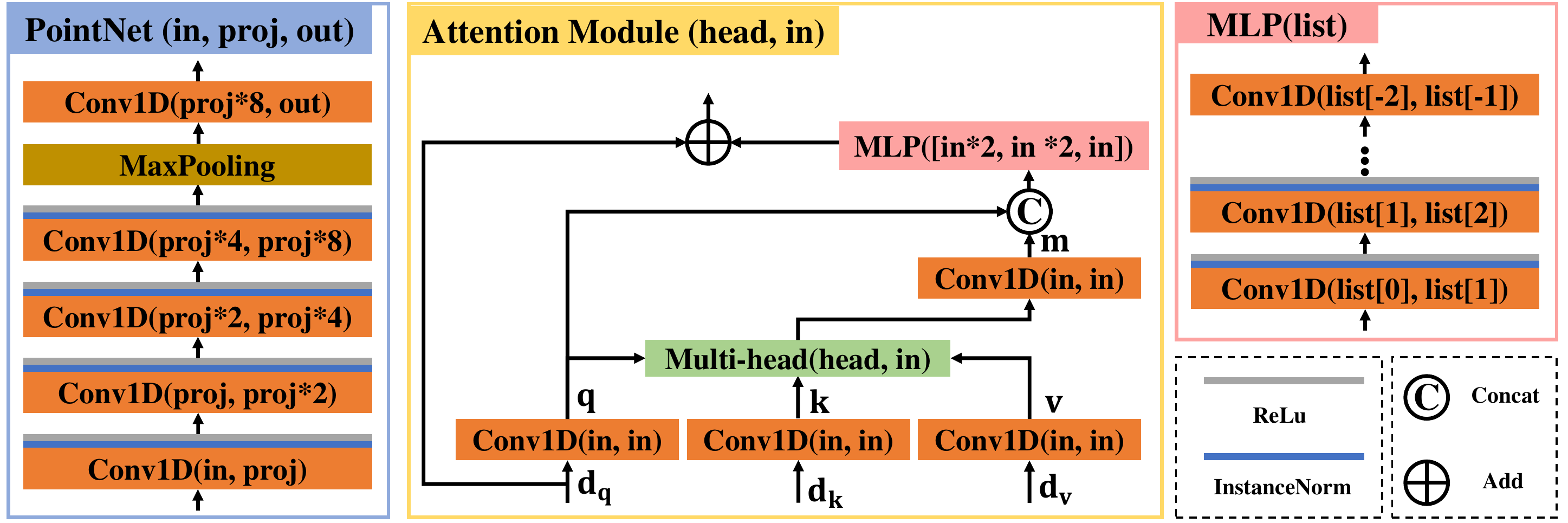}
	\caption{\textbf{Detailed Architecture of Components}. In attention modules, "Multi-head" stands for the multi-head mechanism~\cite{vaswani2017attention}, where $\mathbf{q}$, $\mathbf{k}$ and $\mathbf{v}\in \mathbb{R}^{\mathbf{in}}$ are first reshaped to $(\mathbf{head}, \mathbf{in}/\mathbf{head})$, and attention is then computed separately for each head channel from corresponding $\mathbf{q}$ and $\mathbf{k}$. \textit{Value} $\mathbf{v}$ in each head channel is fused independently according to the attention computed for the same head. The fused \textit{values} with shape \textbf{(head, in/head)} are reshaped back to $\mathbf{(in,1)}$, which is finally projected to message $\mathbf{m}\in \mathbb{R}^{\mathbf{in}}$.}
	\label{fig:arch}
\end{figure*}

\subsection{Loss Functions}
\label{Sec.3_7}
The total loss function $\mathcal{L} = \mathcal{L}_c + \lambda \mathcal{L}_f$ consists of a coarse-level matching loss $\mathcal{L}_c$  and a fine-scale correspondence refinement loss $\mathcal{L}_f$. $\lambda \in \mathbb{R}$ is the hyper-parameter used to balance the two terms.

\subsubsection{Coarse-level Loss for Node Matching}
Following ~\cite{yu2021cofinet}, our coarse-level loss is defined according to the overlap ratios of the vicinities $(\mathcal{V}^{\mathcal{X}^\prime}_i, \mathcal{V}^{\mathcal{Y}^\prime}_j)$ of each node correspondence $(\mathbf{x}^\prime_i, \mathbf{y}^\prime_j)$. Given vicinities $(\mathcal{V}^{\mathcal{X}^\prime}_i, \mathcal{V}^{\mathcal{Y}^\prime}_j)$ of node correspondence $(\mathbf{x}^\prime_i, \mathbf{y}^\prime_j)$, the number of visible points in one vicinity w.r.t. the other vicinity is defined as:
\begin{equation}
n^{j}_{i} = \sum_{\mathbf{x}_u \in \mathcal{V}^{\mathcal{X}^\prime}_i} \mathds{1}(\exists \mathbf{y}_v \in \mathcal{V}^{\mathcal{Y}^\prime}_j s.t. \lVert\mathbf{T}(\mathbf{x}_u) - \mathbf{y}_v\rVert_2 < \tau_p),
\label{eq:nonocc1}
\end{equation}
and
\begin{equation}
n^{i}_{j} = \sum_{\mathbf{y}_v \in \mathcal{V}^{\mathcal{Y}^\prime}_j} \mathds{1}(\exists \mathbf{x}_u \in \mathcal{V}^{\mathcal{X}^\prime}_i s.t. \lVert\mathbf{T}(\mathbf{x}_u) - \mathbf{y}_v\rVert_2 < \tau_p),
\label{eq:nonocc2}
\end{equation}
\noindent for vicinities $\mathcal{V}^{\mathcal{X}^\prime}_i$ and $\mathcal{V}^{\mathcal{Y}^\prime}_j$, respectively, where $\tau_p \in \mathbb{R}$ is the distance threshold for correspondence decision. The overlap ratio between vicinities $(\mathcal{V}^{\mathcal{X}^\prime}_i, \mathcal{V}^{\mathcal{Y}^\prime}_j)$ is further defined as $r^j_i = \frac{1}{2}(\frac{n^{j}_{i}}{|\mathcal{V}^{\mathcal{X}^\prime}_i|} + \frac{n^{i}_{j}}{|\mathcal{V}^{\mathcal{Y}^\prime}_j|})$.

Similar to~\cite{bai2020d3feat,huang2021predator,qin2022geometric}, we use Circle Loss~\cite{sun2020circle}, a variant of Triplet Loss~\cite{schroff2015facenet}, to guide the learning of node descriptors. For a node $\mathbf{x}^\prime_i$ from $\mathcal{X}^\prime$, we sample a positive set $\mathcal{E}^i_p$ composed of nodes $\mathbf{y}^\prime_j$ from $\mathcal{Y}^\prime$ s.t. $\mathbf{T}(\mathcal{V}^{\mathcal{X}^\prime}_i)$ overlaps with $\mathcal{V}^{\mathcal{Y}^\prime}_j$, and a negative set $\mathcal{E}^i_n$ consisting of nodes $\mathbf{y}^\prime_l$ from $\mathcal{Y}^\prime$ s.t. $\mathbf{T}(\mathcal{V}^{\mathcal{X}^\prime}_i)$ and $\mathcal{V}^{\mathcal{Y}^\prime}_l$ share no overlap, where $\mathbf{T}(\mathcal{V}^{\mathcal{X}^\prime}_i)$ denote $\mathcal{V}^{\mathcal{X}^\prime}_i$ transformed by the ground truth transformation $\mathbf{T}$. The loss function on $\mathcal{X}^\prime$ can be defined upon $n$ nodes $\mathbf{x}^\prime_i$ sampled from $\mathcal{X}^\prime$ as:

\begin{equation}
\mathcal{L}^{\mathcal{X}^\prime}_c = \frac{1}{n}\sum_{i=1}^{n} \log\big[1 + \sum_{\mathbf{y}^\prime_j \in \mathcal{E}_p^i}e^{r_i^j \beta^j_p (d^j_i - \Delta_p)} \cdot \sum_{\mathbf{y}^\prime_l \in \mathcal{E}_n^i}e^{\beta^l_n (\Delta_n - d^l_i)}\big],
\label{eq:coarse_loss}
\end{equation}
where $r_i^j$ is the overlap ratio between $\mathcal{V}^{\mathcal{X}^\prime}_i$ and $\mathcal{V}^{\mathcal{Y}^\prime}_j$, and $d_i^j = \lVert \hat{\mathbf{d}}^{\mathcal{X}^\prime}_i - \hat{\mathbf{d}}^{\mathcal{Y}^\prime}_j\rVert_2$ denotes the Euclidean distance of nodes $\mathbf{x}^\prime_i$ and $\mathbf{y}^\prime_j$ in learned feature space. $\Delta_p$ and $\Delta_n$ are the positive and negative margins, which are set to 0.1 and 1.4 in practice, respectively. Furthermore, $\beta^j_p=\gamma(d_i^j - \Delta_p)$ and $\beta^l_n=\gamma(\Delta_n - d_i^l)$ are the weights determined for each sample individually, with the same hyper-parameter $\gamma \in \mathbb{R}$. We can similarly define the loss $\mathcal{L}^{\mathcal{Y}^\prime}_c$ and write the total coarse-level loss as $\mathcal{L}_c = \frac{1}{2}(\mathcal{L}^{\mathcal{X}^\prime}_c + \mathcal{L}^{\mathcal{Y}^\prime}_c)$.

\subsubsection{Fine-level Loss for Correspondence Refinement}
After getting the coarse correspondence set $\mathcal{C}$, we adopt a negative log-likelihood loss~\cite{sarlin2020superglue} to guide the correspondence refinement procedure. For node correspondence $C_l \defeq (\mathbf{x}^\prime_i, \mathbf{y}^\prime_j)\in \mathcal{C}$, as mentioned before, we compute its confidence matrix $\hat{\mathbf{Z}}_l\in \mathbb{R}^{|\mathcal{D}^{\mathcal{X}^\prime}_i + 1|\times|\mathcal{D}^{\mathcal{Y}^\prime}_j + 1|}$ augmented with a slack row and slack column for no correspondence. The ground truth point correspondence set between vicinities $\mathcal{V}^{\mathcal{X}^\prime}_i$ and $ \mathcal{V}^{\mathcal{Y}^\prime}_j$ is denoted as $\mathcal{M}_l$, while the sets of unmatched points in vicinity $\mathcal{V}^{\mathcal{X}^\prime}_i$ and $\mathcal{V}^{\mathcal{Y}^\prime}_j$ are represented as $\mathcal{I}_l$ and $\mathcal{J}_l$, respectively. The ground truth point correspondence set between vicinities $\mathcal{V}^{\mathcal{X}^\prime}_i$ and $ \mathcal{V}^{\mathcal{Y}^\prime}_j$ is defined as:
\begin{equation}
\mathcal{M}_l=\{(\mathbf{x}_u\in \mathcal{V}^{\mathcal{X}^\prime}_i, \mathbf{y}_v\in \mathcal{V}^{\mathcal{Y}^\prime}_j) \big| \lVert\mathbf{T}(\mathbf{x}_u)-\mathbf{y}_v\rVert_2 < \tau_p\}.
\label{eq:vicinity_corr}
\end{equation}
\noindent The set of occluded points in one vicinity w.r.t. the other one is defined as:

\begin{equation}
    \mathcal{I}_l = \{\mathbf{x}_u\in \mathcal{V}^{\mathcal{X}^\prime}_i\big|\nexists \mathbf{y}_v\in \mathcal{V}^{\mathcal{Y}^\prime}_j s.t. \lVert\mathbf{T}(\mathbf{x}_u)-\mathbf{y}_v\rVert_2 < \tau_p\},
\label{eq:occ1}
\end{equation} 
\noindent and 
\begin{equation}
\mathcal{J}_l = \{\mathbf{y}_v\in \mathcal{V}^{\mathcal{Y}^\prime}_j\big| \nexists \mathbf{x}_u\in \mathcal{V}^{\mathcal{X}^\prime}_i s.t. \lVert\mathbf{T}(\mathbf{x}_u)-\mathbf{y}_v\rVert_2 < \tau_p\},
\label{eq:occ2}
\end{equation}
\noindent for vicinities $\mathcal{V}^{\mathcal{X}^\prime}_i$ and $\mathcal{V}^{\mathcal{Y}^\prime}_j$, respectively. 

Finally, the correspondence refinement loss of $C_l$ reads as:

\begin{equation}
\begin{aligned}
\mathcal{L}_f^l = -\sum_{(\mathbf{x}_u, \mathbf{y}_v)\in \mathcal{M}_l} \log \hat{\mathbf{Z}}_l^{u, v} - \sum_{\mathbf{x}_u\in \mathcal{I}_l} \log \hat{\mathbf{Z}}_l^{u, |\mathcal{D}^{\mathcal{Y}^\prime}_j| + 1} \\ - \sum_{\mathbf{y}_v\in \mathcal{J}_l} \log \hat{\mathbf{Z}}_l^{|\mathcal{D}^{\mathcal{X}^\prime}_i| + 1, v}
\end{aligned}
\label{eq:refinement}
\end{equation}

\noindent where $\hat{\mathbf{Z}}_l^{u, v}$ denotes the entry of $\hat{\mathbf{Z}}_l$ on the $u^{th}$ row and $v^{th}$ column. The total loss is averaged across the whole node correspondence set $\mathcal{C}$ as $\mathcal{L}_f = \frac{1}{|\mathcal{C}|}\sum_{l=0}^{|\mathcal{C}|}\mathcal{L}_f^l$.

%% file: sections/4result.tex
\section{Results}
We evaluate \OURS{} on both synthetic object dataset ModelNet40~\cite{wu20153d} and real scene benchmarks, including 3DMatch~\cite{zeng20173dmatch} and 3DLoMatch~\cite{huang2021predator}. RANSAC~\cite{fischler1981random} is leveraged to estimate transformation based on putative correspondences. We further demonstrate our robustness against poor normal estimation in the Appendix by using KITTI~\cite{geiger2012we}. We also compare \OURS{} to the state-of-the-art methods in terms of inference speed in the Appendix. Qualitative results can be found in Fig.~\ref{fig:visualization}. We also illustrate failed cases from 3DLoMatch in Fig.~\ref{fig:failed_3dmatch}. More qualitative results on ModelNet40, 3DMatch, and 3DLoMatch are provided in the Appendix.

\subsection{Implementation Details}
\label{sec:implementation}
\subsubsection{Detailed Architecture}
\label{sec:architecture}
The detailed architecture of each component leveraged in \OURS{} can be found in Fig.~\ref{fig:arch}. PointNets~\cite{qi2017pointnet} $\Psi_g$ and $\Psi_s$ are two individual models with the same architecture~(input dimension $\mathbf{in}=4$, project dimension $\mathbf{proj}=64$ and output dimension $\mathbf{out}=256$), as shown in the leftmost column in Fig.~\ref{fig:arch}. Each attention block has an intra-frame module and an inter-frame module, both with the architecture of the "Attention Module" shown in Fig.~\ref{fig:arch}. Differently, for intra-frame modules, $\mathbf{d}_q$, $\mathbf{d}_k$ and $\mathbf{d}_v$ are all from the same frame, while in inter-frame modules, $\mathbf{d}_k$ and $\mathbf{d}_v$ are from the opposite frame. $\mathbf{MLP_c}$ and $\mathbf{MLP_f}$ in Fig.~\ref{fig:c2f} have the same MLP architecture shown in the rightmost column of Fig.~\ref{fig:arch}, with a input dimension list of [256, 128, 64, 32]. 

\subsubsection{Training and Testing} \OURS{} is implemented with PyTorch~\cite{paszke2019pytorch} and trained end-to-end on a single NVIDIA RTX 3090 with 24G memory, where the batch size is set to 2 for 3DMatch/3DLoMatch~\cite{zeng20173dmatch,huang2021predator} and 16 for ModelNet40~\cite{wu20153d}. Notably, it could also be trained on a GPU with 11G memory, e.g., NVIDIA GTX 1080Ti. We train for 150 epochs on ModelNet40 and for 20 epochs on 3DMatch/3DLoMatch, both with $\lambda=1$ to balance different loss functions. We leverage an Adam optimizer~\cite{kingma2014adam} with an initial learning rate of 1e-4, which is exponentially decayed by 0.05 after each epoch. On ModelNet40, we sparsely sample $N^\prime=M^\prime=256$ nodes from each point cloud pair, with a radius $r=0.2m$ to construct support areas, within which the number of points is truncated to 64. On 3DMatch/3DLoMatch, $N^\prime$ and $M^\prime$ are both set to 512, with $r=0.3m$ and 512 points within each support area. Besides, the number of points in vicinity $\mathcal{V}$ is truncated to 32 and 128 on ModelNet40 and 3DMatch/3DLoMatch respectively. On both datasets, the dimension of intermediate descriptors $\mathbf{g}$, $\mathbf{s}$ and $\widetilde{\mathbf{d}}$ is set to 256, while that of descriptors $\hat{\mathbf{d}}$, from which correspondences are hierarchically extracted, is set to 32. The number of neighbor points used for feature interpolation is set to $k=3$. We use 100 iterations for Sinkhorn~\cite{sinkhorn1967concerning} algorithm. The number of attention blocks is set to \textit{K}=6, and the attention mechanism is implemented with 4 heads. During training, 256 node pairs that overlap under ground truth transformation are sampled as the node correspondence set $\mathcal{C}$. During testing, 256 node correspondences with the highest similarity scores are selected for the consecutive refinement.

\begin{figure*}[h]
	\centering
	\includegraphics[width=0.95\linewidth]{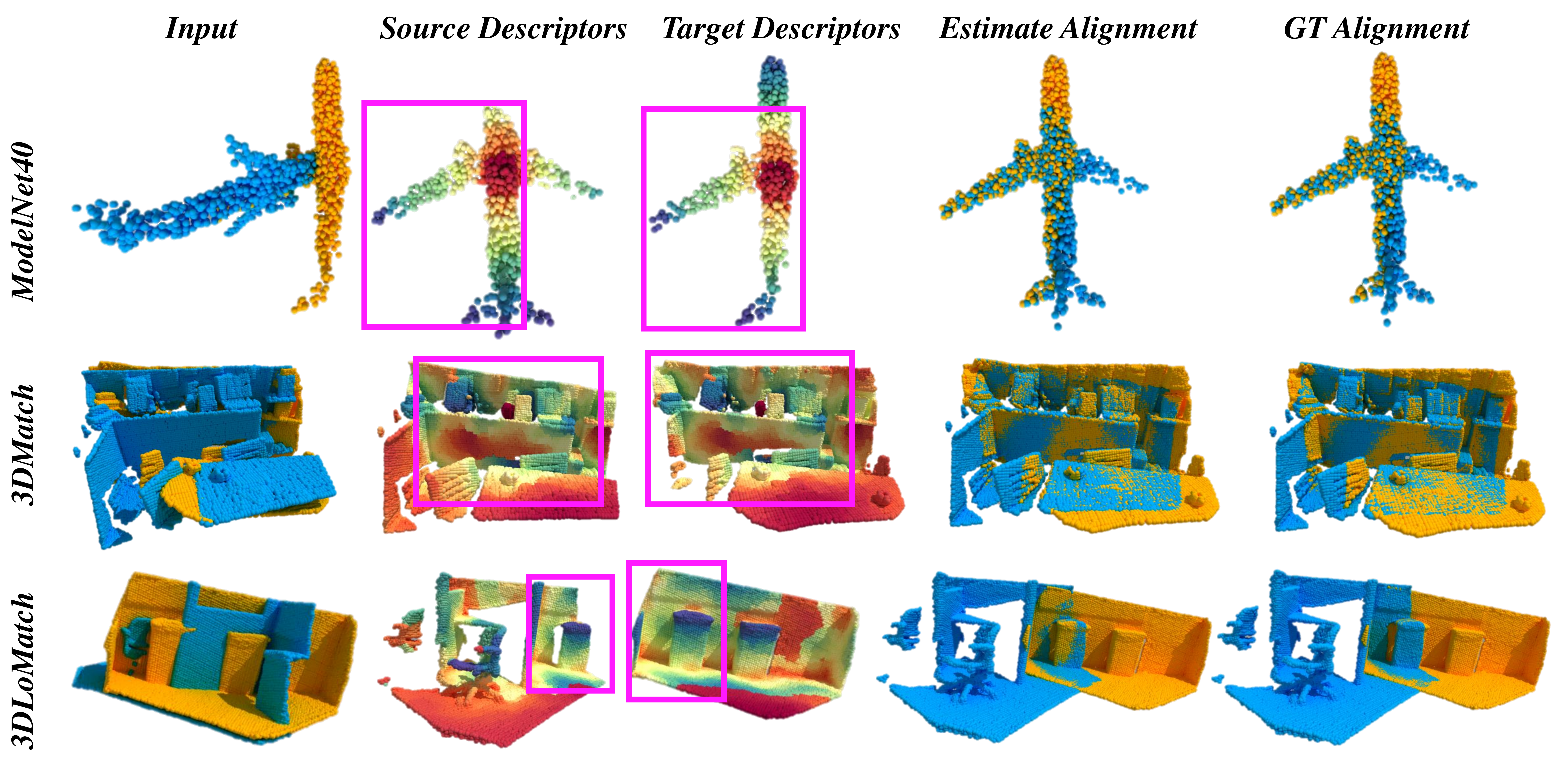}
	\caption{\textbf{Qualitative Results.} We use t-SNE~\cite{van2008visualizing} to visualize the learned descriptors of source and target point clouds. In the rectangles, we roughly demonstrate the overlap regions.}
	\label{fig:visualization}
\end{figure*}

\subsection{Synthetic Object Dataset: ModelNet40}
\subsubsection{Dataset} ModelNet40~\cite{wu20153d} consists of 12,311 CAD models of objects from 40 different categories. We follow the setting of~\cite{wang2019prnet}, where 9,833 shapes are used for training, and the rest 2,468 for testing. For each model, 1,024 points are randomly sampled from its surface. For simulating the partial overlap from scanning, 768 points nearest to a randomly selected viewpoint in the space are resampled from the 1,024 points, which serves as the input point cloud. Following~\cite{pan2021robust}, instead of using the ground truth normals, we estimate them using Open3D~\cite{zhou2018open3d}.

\begin{table*}[h!]
\caption{\textbf{Results on ModelNet40.} Best
performance is highlighted in bold while the second best is marked with an underline. In "Unseen", 20 categories are used for training and the rest 20 for testing. In "Noise", all the categories are split into training and testing. Gaussian noise sampled from $\mathcal{N}(0, 0.01)$ and clipped to [-0.05, 0.05] is added to individual points in both training and testing. In "[0, 45\degree]", rotations along each axis are randomly sampled from [0, 45\degree] and translations are sampled from [-0.5, 0.5]. Rotations are enlarged to 180\degree in "[0, 180\degree]".
}
\resizebox{\textwidth}{!}{
\begin{tabular}{lccccccc|cccccc}
\toprule
\multirow{3}{*}{\textBF{Methods}} & & &\multicolumn{4}{c}{\textBF{Unseen}} & & &\multicolumn{4}{c}{\textBF{Noise}} & \\
& & &[0, 45\degree] & & &[0, 180\degree] & & &[0, 45\degree] & & &[0, 180\degree] &\\
&\#dim &\textit{RRE} $\downarrow$&\textit{RTE} $\downarrow$ &\textit{RMSE} $\downarrow$ &\textit{RRE} $\downarrow$ &\textit{RTE} $\downarrow$ &\textit{RMSE} $\downarrow$ &\textit{RRE} $\downarrow$ &\textit{RTE} $\downarrow$ &\textit{RMSE} $\downarrow$ &\textit{RRE} $\downarrow$ &\textit{RTE} $\downarrow$ &\textit{RMSE} $\downarrow$\\
\hline \hline
PRNet~\cite{wang2019prnet} &1024 &3.19\degree &0.028 &0.036 &91.94\degree &0.297 &0.545 &4.37\degree &0.034 &0.045 &95.80\degree &0.319 &0.542 \\

IDAM~\cite{li2020iterative} &32 &0.86\degree &0.005 &0.007 &16.17\degree &0.073 &0.106 &9.60\degree &0.052 &0.084 &71.06\degree &0.217 &0.430\\

RPM~\cite{fu2021robust} &1024 &0.34\degree &0.004 &0.004 &8.78\degree &0.076 &0.084 &2.21\degree &0.013 &0.018 &23.58\degree &0.111 &0.156 \\

DCP~\cite{wang2019deep} &1024 &11.92\degree &0.076 &0.119 &67.39\degree &0.170 &0.410 &9.33\degree &0.070 &0.097 &73.61\degree &0.185 &0.441 \\ 

DeepGMR~\cite{yuan2020deepgmr} &128 &17.45\degree &0.074 &0.130 &49.23\degree &0.219 &0.349 &16.96\degree &0.068 &0.120 &68.68\degree &0.248 &0.419\\

RPMNet~\cite{yew2020rpm} &96 &0.60\degree &0.004 &0.005 &16.91\degree &0.079 &0.127 &3.52\degree &0.214 &0.029 &37.82\degree &0.132 &0.250\\

GMCNet~\cite{pan2021robust} &128 &\underline{0.026\degree} &\underline{0.0002} &\underline{0.0002} &\textBF{0.39\degree} &\textBF{0.002} &\textBF{0.003} &\textBF{0.94\degree} &\underline{0.007} &\textBF{0.008} &{18.13\degree} &{0.093} &{0.132}\\

Predator~\cite{huang2021predator} &96 &1.32\degree &0.009 &0.012 &11.59\degree &\underline{0.032} &0.058 &3.33\degree &0.018 &0.025 &40.64\degree &0.110 &0.207\\

CoFiNet~\cite{yu2021cofinet} &32 &2.30\degree &0.027 &0.033 &6.55\degree &0.033 &\underline{0.056} &3.06\degree &0.017 &0.027 &\underline{14.33}\degree &\underline{0.034} &\underline{0.091}\\

\OURS{}~&32  &\textBF{0.004\degree} &\textBF{\textless 0.0001} &\textBF{\textless 0.0001} &\underline{0.41}\degree &\textBF{0.002} &\textBF{0.003} &\underline{1.15}\degree &\textBF{0.006} &\underline{0.009} &\textBF{5.99\degree} &\textBF{0.008} &\textBF{0.029} \\



\bottomrule
\end{tabular}}

\label{tab:object}
\end{table*}

\begin{figure*}[h!]

\centering
\includegraphics[width=\linewidth]{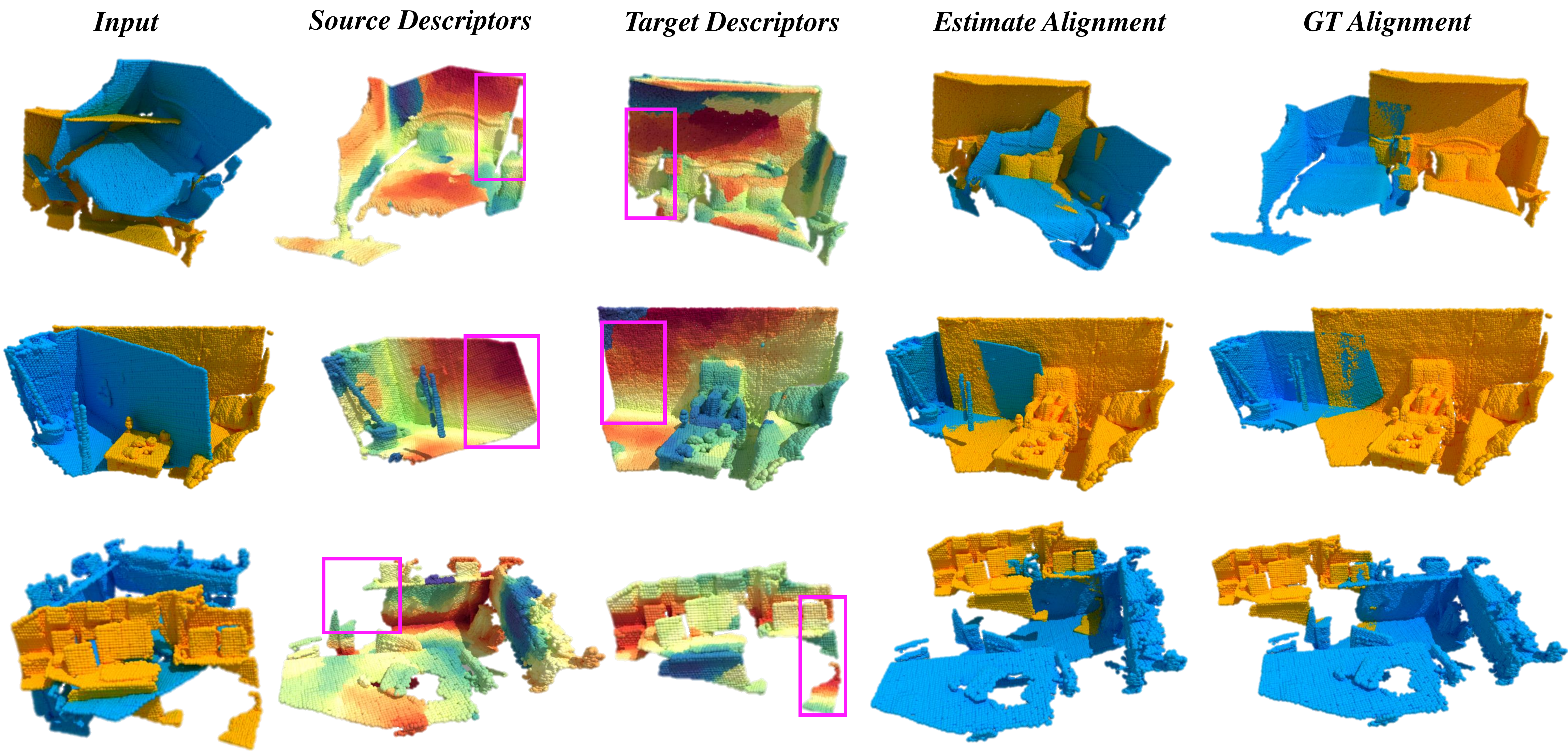}
\caption{\textbf{Failed cases on 3DLoMatch}. We use t-SNE~\cite{van2008visualizing} to visualize the learned descriptors of source and target point clouds. In the rectangles, we roughly demonstrate the overlap regions. The failed cases have reasonable descriptors but extremely limited overlap.}
\label{fig:failed_3dmatch}

\end{figure*}
\begin{table}[h!]
\centering
\caption{\textbf{Comparisons to the State-of-the-Art on 3DMatch and 3DLoMatch.} Best
performance is highlighted in bold while the second best is marked with an underline. In column "Rotated"\textsuperscript{\ref{note:rotated}}, every point cloud pair is evaluated with \# Samples=5,000\textsuperscript{\ref{note:lepard}}~(in Tab.~\ref{tab:scene} and Tab.~\ref{tab:scene_rotated}), and each point cloud is rotated individually with random rotations up to 360\degree along each axis. Our method significantly outperforms state-of-the-art methods on the rotated benchmarks.}
\resizebox{0.48\textwidth}{!}{
\begin{tabular}{lcc|cc}
\toprule
&\multicolumn{2}{c}{\textBF{3DMatch}} &\multicolumn{2}{c}{\textBF{3DLoMatch}}\\
\# Samples &Origin &Rotated &Origin &Rotated\\
\midrule
\midrule
& &\multicolumn{2}{c}{\textit{Inlier Ratio}(\%) $\uparrow$}&\\
\midrule
3DSN~\cite{gojcic2019perfect} &36.0 &- &11.4 &-\\
FCGF~\cite{choy2019fully} &56.8 &49.3 &21.4 &17.3\\
D3Feat~\cite{bai2020d3feat} &39.0 &37.7 &13.2 &12.1\\
SpinNet~\cite{ao2021spinnet} &48.5 &48.7 &25.7 &\underline{25.7}\\
Predator~\cite{huang2021predator} &58.0 &{52.8}  &26.7 &22.4\\
YOHO~\cite{wang2021you} &\underline{64.4} &\underline{64.1} &25.9 &23.2\\
CoFiNet~\cite{yu2021cofinet} &49.8 &46.8 &24.4  &21.5\\
Lepard~\cite{li2022lepard} &58.6 &53.7 &\underline{28.4} &{24.4}\\
\OURS{}  &\textBF{68.4} &\textBF{68.5} &\textBF{32.1} &\textBF{32.1}\\
\midrule
& &\multicolumn{2}{c}{\textit{Feature Matching Recall}(\%) $\uparrow$} &\\
\midrule
3DSN~\cite{gojcic2019perfect}  &95.0 &-  &63.6 &-\\
FCGF~\cite{choy2019fully}  &97.4 &96.9  &76.6 &73.3 \\
D3Feat~\cite{bai2020d3feat} &95.6  &94.7  &67.3 &63.9\\
SpinNet~\cite{ao2021spinnet}  &97.4 &{97.4}  &75.5 &75.2 \\
Predator~\cite{huang2021predator}  &96.6 &96.2 &78.6  &73.7 \\
YOHO~\cite{wang2021you} &\textBF{98.2} &\underline{97.8} &79.4 &77.8\\
CoFiNet~\cite{yu2021cofinet}  &\underline{98.1} &{97.4} &\underline{83.1} &{78.6}\\
Lepard~\cite{li2022lepard} &98.0 &97.4 &\underline{83.1} &\underline{79.5}\\
\OURS{} &97.9 &\textBF{98.2}  &\textBF{85.1} &\textBF{84.5}\\
\midrule
& &\multicolumn{2}{c}{\textit{Registration Recall}(\%) $\uparrow$} &\\
\midrule
3DSN~\cite{gojcic2019perfect}  &78.4 &-  &33.0 &-\\
FCGF~\cite{choy2019fully} &85.1 &90.3 &40.1 &58.6\\
D3Feat~\cite{bai2020d3feat}  &81.6 &91.3 &37.2 &55.3 \\
SpinNet~\cite{ao2021spinnet}  &88.8 &\textBF{93.2}  &58.2 &61.8\\
Predator~\cite{huang2021predator}  &{89.0} &92.0  &59.8 &58.6\\
YOHO~\cite{wang2021you} &\underline{90.8} &92.5 &65.2 &\underline{66.8}\\
CoFiNet~\cite{yu2021cofinet}  &{89.3} &92.0  &\textBF{67.5} &{62.5}\\
Lepard~\cite{li2022lepard} &\textBF{92.7} &84.9 &\underline{65.4} &49.0\\
\OURS{}~ &{89.3} &\underline{93.0}  &{65.1} &\textBF{66.9}\\
\bottomrule
\end{tabular}}

\label{tab:compare}

\end{table}

\subsubsection{Metrics}
We use 3 widely-adopted metrics~\cite{pan2021robust}: (1)~\textit{Relative Rotation Error}~(\textbf{RRE}) that evaluates the error between estimated and ground truth rotation matrices; (2)~\textit{Relative Translation Error}~(\textbf{RTE}) that measures the error between estimated and ground truth translation vectors; (3)~\textit{Root-Mean-Square Error}~(\textbf{RMSE}) which calculates the residual error between correspondences from the same point cloud, separately transformed by the estimated and ground truth transformation. Please refer to the Appendix for the detailed definition.

\subsubsection{Comparisons to the State-of-the-Art}
We compare \OURS{} with 9 state-of-the-art baselines, including 7 direct registration methods and 2 correspondence-based approaches~(Predator~\cite{huang2021predator} and CoFiNet~\cite{yu2021cofinet}). The detailed results are shown in Tab.~\ref{tab:object}. From the second column that lists the dimension of descriptors used for correspondence search, it can be noticed that \OURS{} uses the most compact descriptors among all the methods. On the "Unseen" setting, \OURS{} surpasses all the other methods with rotations in the range of [0, 45\degree]. With a maximum rotation of 180\degree, it achieves on-par performance with GMCNet~\cite{pan2021robust} and outperforms others. When Gaussian noise is added, although \OURS{} stays comparable with GMCNet~\cite{pan2021robust} with rotations in [0, 45\degree], it outperforms all the baselines on all the metrics by a large margin with rotations enlarged to 180\degree. Notably, all the methods except for \OURS{} degenerate significantly, which shows the superiority of the inherent rotational invariance of \OURS{}. Although direct registration methods are specifically tuned with good performance on object-level data as pointed out in~\cite{huang2021predator}, \OURS{} could compete with them and even performs significantly better than them on data with Gaussian noise and large rotations. Moreover, \OURS{} also achieves the state-of-the-art performance on scene-level benchmarks~\cite{zeng20173dmatch,huang2021predator}, while most direct registration methods fail to work there according to~\cite{huang2021predator}.

\subsection{Real Scene Benchmarks: 3DMatch and 3DLoMatch}
\subsubsection{Datasets}
3DMatch~\cite{zeng20173dmatch} collects 62 scenes, where 46 scenes are used for training, 8 for validation, and the rest 8 for testing. We use the processed data and split in~\cite{huang2021predator}, and evaluate \OURS{} on both 3DMatch~\cite{zeng20173dmatch}~(\textgreater 30\% overlap) and 3DLoMatch~\cite{huang2021predator}~(10\% $\sim$ 30\% overlap) protocols. Additionally, we also follow~\cite{deng2018ppf,ao2021spinnet} to test on benchmarks with enlarged rotations to demonstrate the superiority of the inherent rotational invariance of our descriptors.

\subsubsection{Metrics}
We follow~\cite{huang2021predator,yu2021cofinet} and use 3 metrics for evaluation: (1)~\textit{Inlier Ratio}~(\textbf{IR}), which is the fraction of putative correspondences whose residual error is lower than a threshold $\tau_2=0.1m$ under the ground truth transformation, and (2)~\textit{Feature Matching Recall}~(\textbf{FMR}) that counts the fraction of point cloud pairs whose \textbf{Inlier Ratio} is larger than a threshold $\tau_1=5\%$, and (3)~\textit{Registration recall}~(\textbf{RR}) that stands for the fraction of point cloud pairs whose \textbf{RMSE} between the estimated and ground truth transformation is smaller than a threshold $\tau_3=0.2m$. \footnote{\scriptsize{Instead of strictly following the criterion, we follow~\cite{huang2021predator,yu2021cofinet} to calculate \textbf{RR} according to pre-computed correspondences defined on original 3DMatch/3DLoMatch.}} Please refer to the Appendix for details.

\subsubsection{Comparisons to the State-of-the-Art}
In Tab.~\ref{tab:scene}, we compare \OURS{} with 8 baseline methods. Specifically, 3DSN~\cite{gojcic2019perfect}, SpinNet~\cite{ao2021spinnet}, and YOHO~\cite{wang2021you} are rotation-invariant approaches without global awareness. Predator~\cite{huang2021predator}, CoFiNet~\cite{yu2021cofinet}, and Lepard\footnote{\scriptsize{We use the criterion in ~\cite{huang2021predator} and~\cite{yu2021cofinet} to evaluate~\cite{li2022lepard} and use all the correspondences without sampling following~\cite{li2022lepard}}\label{note:lepard}}~\cite{li2022lepard} are globally-aware algorithms that are variant to rotations. We validate our method on both original and rotated benchmarks.\footnote{\scriptsize{On rotated data, \textbf{RR} is calculated with \textbf{RMSE}\textless 0.2m, which is different to \textbf{RR} on original data.}\label{note:rotated}} For \textbf{IR}, \OURS{} significantly outperforms all the baselines on original 3DMatch and 3DLoMatch, which indicates \OURS{} learns more distinctive descriptors and extracts more reliable correspondences. When the benchmarks are further rotated, our superiority over others becomes more significant, which demonstrates the advantage of our rotational invariance by design. Notably, with larger rotations, only the performance of SpinNet~\cite{ao2021spinnet}, YOHO~\cite{wang2021you}, and \OURS{} remains stable, which further proves the superiority of inherent rotational invariance over the learned one. For \textbf{FMR}, we perform the best on rotated data. When rotations are enlarged, especially on 3DLoMatch, the performance of all the methods except for \OURS{} and SpinNet~\cite{ao2021spinnet} drops sharply. The performance drop of YOHO further demonstrates the aforementioned drawback of achieving rotational invariance via equivariance. Moreover, due to the lack of global awareness, SpinNet~\cite{ao2021spinnet} falls behind Predator\cite{huang2021predator}, CoFiNet~\cite{yu2021cofinet}, Lepard~\cite{li2022lepard}, and \OURS{} in terms of \textbf{FMR}, which supports the significance of being globally-aware. Finally, for \textbf{RR}, we perform on-par with CoFiNet~\cite{yu2021cofinet} and Lepard~\cite{li2022lepard} on original datasets, but again show our excellence when rotations are enlarged.

\subsubsection{Detailed Results with Different Numbers of Samples}
\begin{figure}[h!]
\centering
\includegraphics[width=\linewidth]{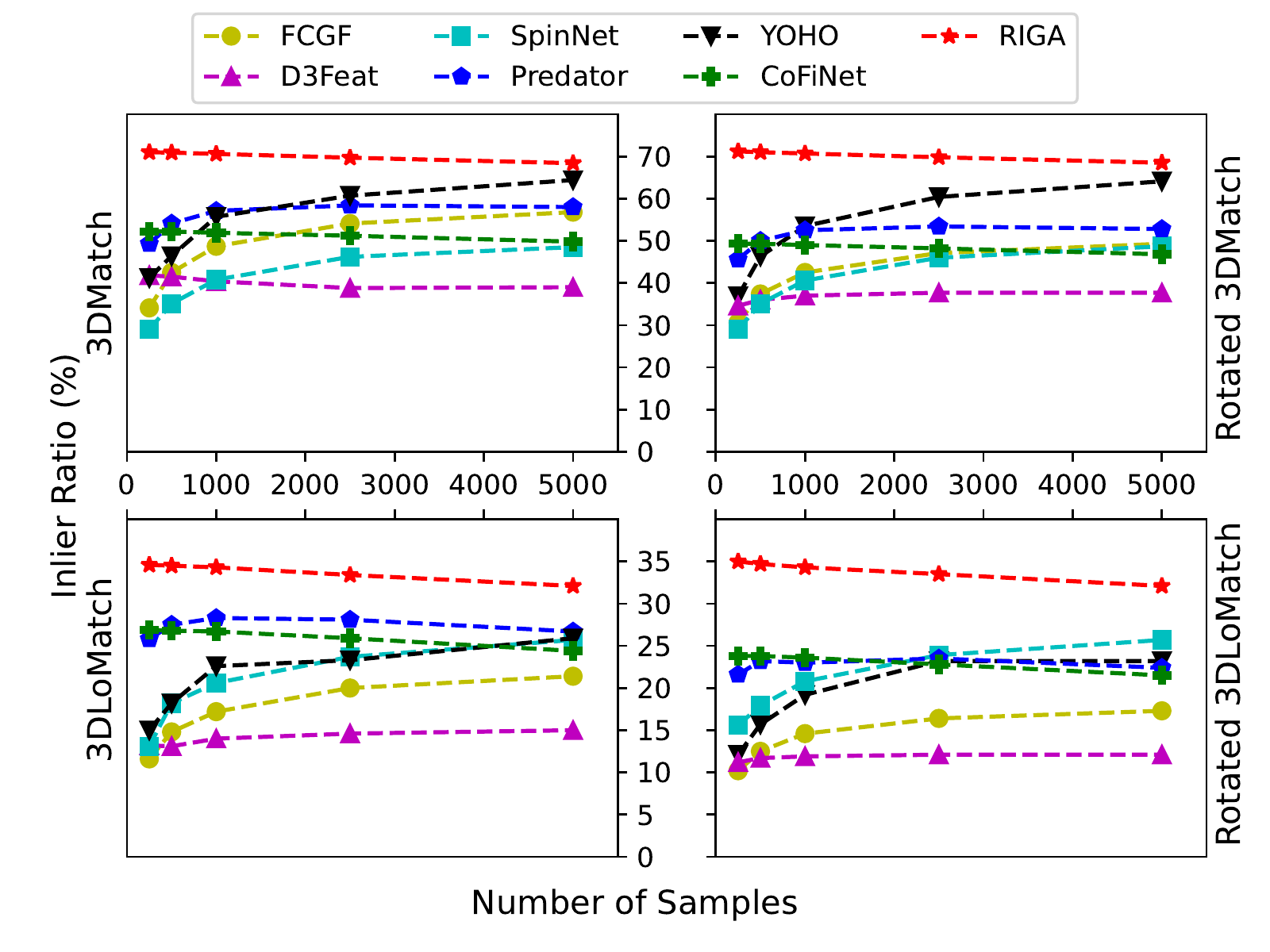}
\caption{\textbf{Inlier Ratio~(IR) with Different Numbers of Samples.} \OURS{} achieves the best performance on all the datasets. Notably, the performance of \OURS{} increases when the number of sampled correspondences decreases, which further demonstrates the superiority of our coarse-to-fine mechanism for correspondence extraction.}
\label{fig:ir_sample}

\end{figure}

In Tab.~\ref{tab:scene}, Tab.~\ref{tab:scene_rotated} and Fig.~\ref{fig:ir_sample}, we follow~\cite{huang2021predator,yu2021cofinet} to show the performance with different numbers of sampled points/correspondences. The \textBF{IR} of CoFiNet~\cite{yu2021cofinet} and \OURS{} increases when the number of samples decreases. This is because methods with the coarse-to-fine matching mechanism implicitly consider all the potential correspondences and sample the most confident ones for registration, while methods relying on uniform sub-sampling or keypoint detection only extract correspondences from sparsely-sampled nodes, whose repeatability is hard to guarantee especially with fewer samples. When the sample number is decreased from 5,000 to 250, all the other metrics of CoFiNet and \OURS{} remain stable, while those of the others usually drop significantly, which further proves the excellence of the coarse-to-fine mechanism against fewer samples.

\begin{table}[h!]
\centering
\caption{\textbf{Quantitative Results on 3DMatch and 3DLoMatch with Different Numbers of Samples.} Best
performance is highlighted in bold while the second best is marked with an underline. \# Samples is the number of sampled points or correspondences, following~\cite{huang2021predator} and~\cite{yu2021cofinet}, respectively.}
\resizebox{0.48\textwidth}{!}{
\begin{tabular}{lccccc|ccccc}
\toprule
 &\multicolumn{3}{c}{\textBF{3DMatch}}& &  &\multicolumn{3}{c}{\textBF{3DLoMatch}}& &\\
\# Samples &5000 &2500 &1000 &500 &250 &5000 &2500 &1000 &500 &250 \\
\midrule
\midrule
&\multicolumn{10}{c}{\textit{Inlier Ratio}(\%) $\uparrow$}\\
\midrule
3DSN~\cite{gojcic2019perfect} &36.0 &32.5 &26.4 &21.5 &16.4 &11.4 &10.1 &8.0 &6.4 &4.8 \\
FCGF~\cite{choy2019fully} &{56.8} &{54.1} &48.7 &42.5 &34.1 &21.4 &20.0 &17.2 &14.8 &11.6\\
D3Feat~\cite{bai2020d3feat} &39.0 &38.8 &40.4 &41.5 &41.8 &13.2 &13.1 &14.0 &14.6 &15.0\\
SpinNet~\cite{ao2021spinnet} &48.5 &46.2 &40.8 &35.1 &29.0 &25.7 &23.7 &20.6 &18.2 &13.1\\
Predator~\cite{huang2021predator} &58.0 &58.4 &\underline{57.1} &\underline{54.1} &{49.3}  &\underline{26.7} &\underline{28.1} &\underline{28.3} &\underline{27.5} &{25.8}\\
YOHO~\cite{wang2021you} &\underline{64.4} &\underline{60.7} &55.7 &46.4 &41.2 &25.9 &23.3 &22.6 &18.2 &15.0 \\
CoFiNet~\cite{yu2021cofinet} &49.8 &51.2 &{51.9} &{52.2} &\underline{52.2} &{24.4} &{25.9} &{26.7} &{26.8} &\underline{26.9}\\
\OURS{}~  &\textBF{68.4} &\textBF{69.7} &\textBF{70.6} &\textBF{70.9} &\textBF{71.0} &\textBF{32.1}  &\textBF{33.4} &\textBF{34.3} &\textBF{34.5} &\textBF{34.6}\\
\midrule
&\multicolumn{10}{c}{\textit{Feature Matching Recall}(\%) $\uparrow$}\\
\midrule
3DSN~\cite{gojcic2019perfect} &95.0 &94.3 &92.9 & 90.1 &82.9 &63.6 &61.7 &53.6 &45.2 &34.2\\
FCGF~\cite{choy2019fully} &{97.4} &{97.3} &{97.0} &{96.7} &{96.6} &76.6 &75.4 &74.2 &71.7 &67.3\\
D3Feat~\cite{bai2020d3feat} &95.6 &95.4 &94.5 &94.1 &93.1 &67.3 &66.7 &67.0 &66.7 &66.5\\
SpinNet~\cite{ao2021spinnet} &97.4 &97.0 &96.4 &96.7 &94.8 &75.5 &75.1 &74.2 &69.0 &62.7\\
Predator~\cite{huang2021predator} &96.6 &96.6 &96.5 &96.3 &96.5 &{78.6} &{77.4} &{76.3} &{75.7} &{75.3}\\
YOHO~\cite{wang2021you} &\textBF{98.2} &\text97.6 &97.5 &\underline{97.7} &96.0 &79.4 &78.1 &76.3 &73.8 &69.1\\
CoFiNet~\cite{yu2021cofinet} &\underline{98.1} &\textBF{98.3} &\textBF{98.1} &\textBF{98.2} &\textBF{98.3} &\underline{83.1} &\underline{83.5} &\underline{83.3} &\underline{83.1} &\underline{82.6}\\
\OURS{}~ &97.9 &\underline{97.8} &\underline{97.7} &\underline{97.7} &\underline{97.6} &\textBF{85.1} &\textBF{85.0} &\textBF{85.1} &\textBF{84.3} &\textBF{85.1}\\
\midrule
&\multicolumn{10}{c}{\textit{Registration Recall}(\%) $\uparrow$}\\
\midrule
3DSN~\cite{gojcic2019perfect} &78.4 &76.2& 71.4 & 67.6 & 50.8 & 33.0 & 29.0 & 23.3 & 17.0 & 11.0 \\
FCGF~\cite{choy2019fully} &85.1 &84.7 &83.3 &81.6 &71.4 &40.1 & 41.7 & 38.2 & 35.4 & 26.8\\
D3Feat~\cite{bai2020d3feat} &81.6 &84.5 &83.4 &82.4 &77.9 &37.2 &42.7 &46.9 &43.8 &39.1\\
SpinNet~\cite{ao2021spinnet} &88.8 &88.0 &84.5 &79.0 &69.2 &58.2 &56.7 &49.8 &41.0 &26.7\\
Predator~\cite{huang2021predator}&89.0 &\underline{89.9} &\textBF{90.6} &{88.5} &{86.6} &{59.8} &{61.2} &{62.4} &{60.8} &{58.1}\\
YOHO~\cite{wang2021you}&\textBF{90.8} &\textBF{90.3} &\underline{89.1} &\underline{88.6} &84.5 &\underline{65.2} &\underline{65.5} &63.2 &56.5 &48.0\\
CoFiNet~\cite{yu2021cofinet} &\underline{89.3} &{88.9} &{88.4} &{87.4} &\underline{87.0} &\textBF{67.5} &\textBF{66.2} &\underline{64.2} &\underline{63.1} &\underline{61.0}\\
\OURS{}~ &\underline{89.3} &88.4 &\underline{89.1} &\textBF{89.0} &\textBF{87.7}&{65.1} &{64.7} &\textBF{64.5} &\textBF{64.1} &\textBF{61.8}\\
\bottomrule
\end{tabular}}

\label{tab:scene}

\end{table}

\begin{table}[h!]
\centering

\caption{\textbf{Quantitative results on Rotated 3DMatch and 3DLoMatch with Different Numbers of Samples.} Best
performance is highlighted in bold while the second best is marked with an underline. Each point cloud is rotated individually with random rotations up to 360\degree along each axis.}
\resizebox{0.48\textwidth}{!}{
\begin{tabular}{lccccc|ccccc}
\toprule
 & &\multicolumn{3}{c}{3DMatch}& &  &\multicolumn{3}{c}{3DLoMatch}&\\
\# Samples  &5000 &2500 &1000 &500 &250 &5000 &2500 &1000 &500 &250 \\
\midrule
\midrule
&\multicolumn{10}{c}{\textit{Inlier Ratio}(\%) $\uparrow$}\\
\midrule
FCGF~\cite{choy2019fully} &49.3 &47.1 &42.5 &37.4 &30.6 &17.3 &16.4 &14.6 &12.5 &10.2\\
D3Feat~\cite{bai2020d3feat} &37.7 &37.7 &37.0 &36.0 &34.6 &12.1 &12.1 &11.9 &11.7 &11.2\\
SpinNet~\cite{ao2021spinnet} &48.7 &46.0 &40.6 &35.1 &29.0 &\underline{25.7} &\underline{23.9} &20.8 &17.9 &15.6\\
Predator~\cite{huang2021predator} &52.8 &53.4 &52.5 &\underline{50.0} &45.6 &22.4 &23.5 &23.0 &23.2 &21.6\\
YOHO~\cite{wang2021you} &\underline{64.1} &\underline{60.4} &\underline{53.5} &46.3 &36.9 &23.2 &23.2 &19.2 &15.7 &12.1 \\
CoFiNet~\cite{yu2021cofinet} &46.8 &48.2 &49.0 &49.3 &\underline{49.3} &21.5 &22.8 &\underline{23.6} &\underline{23.8} &\underline{23.8}\\
\OURS{} &\textBF{68.5} &\textBF{69.8} &\textBF{70.7} &\textBF{71.0} &\textBF{71.2} &\textBF{32.1} &\textBF{33.5} &\textBF{34.3} &\textBF{34.7} &\textBF{35.0}\\

\midrule
&\multicolumn{10}{c}{\textit{Feature Matching Recall}(\%) $\uparrow$}\\
\midrule
FCGF~\cite{choy2019fully} &96.9 &96.9 &96.2 &95.9 &94.5 &73.3 &73.4 &71.0 &68.8 &64.5\\
D3Feat~\cite{bai2020d3feat} &94.7 &95.1 &94.3 &93.8 &92.3 &63.9 &64.6 &63.0 &62.1 &59.6\\
SpinNet~\cite{ao2021spinnet} &97.4 &97.4 &96.7 &96.5 &94.1 &75.2 &74.9 &72.6 &69.2 &61.8\\
Predator~\cite{huang2021predator} &96.2 &96.2 &96.6 &96.0 &96.0 &73.7 &74.2 &75.0 &74.8 &73.5\\
YOHO~\cite{wang2021you} &\underline{97.8} &\underline{97.8} &\underline{97.4} &\underline{97.6} &96.4 &77.8 &77.8 &76.3 &73.9 &67.3\\
CoFiNet~\cite{yu2021cofinet}&97.4 &97.4 &97.2 &97.2 &\underline{97.3} &\underline{78.6} &\underline{78.8} &\underline{79.2} &\underline{78.9} &\underline{79.2}\\
\OURS{} &\textBF{98.2} &\textBF{98.2} &\textBF{98.2} &\textBF{98.0} &\textBF{98.1} &\textBF{84.5} &\textBF{84.6} &\textBF{84.5} &\textBF{84.2} &\textBF{84.4}\\
\midrule
&\multicolumn{10}{c}{\textit{Registration Recall}(\%) $\uparrow$}\\
\midrule
FCGF~\cite{choy2019fully} &90.3 &91.2 &90.4 &87.8 &83.3 &58.6 &58.7 &54.7 &44.8 &34.7\\
D3Feat~\cite{bai2020d3feat} &91.3 &90.3 &88.4 &85.2 &80.8 &55.3 &53.5 &47.9 &43.6 &33.5\\
SpinNet~\cite{ao2021spinnet} &\textBF{93.2} &\textBF{93.2} &91.1 &87.4 &77.0 &61.8 &59.1 &53.1 &44.1 &30.7\\
Predator~\cite{huang2021predator} &92.0 &92.8 &92.0 &\textBF{92.2} &89.5 &58.6 &59.5 &60.4 &58.6 &55.8\\
YOHO~\cite{wang2021you} &92.5 &92.3 &\underline{92.4} &90.2 &87.4 &\underline{66.8} &\underline{67.1} &\underline{64.5} &58.2 &44.8\\
CoFiNet~\cite{yu2021cofinet} &92.0 &91.4 &91.0 &90.3 &\underline{89.6} &62.5 &60.9 &60.9 &\underline{59.9} &\underline{56.5}\\
\OURS{} &\underline{93.0} &\underline{93.0} &\textBF{92.6} &\underline{91.8} &\textBF{92.3} &\textBF{66.9} &\textBF{67.6} &\textBF{67.0} &\textBF{66.5} &\textBF{66.2}\\
\bottomrule
\end{tabular}}
\label{tab:scene_rotated}
\end{table}

\subsubsection{Scene-wise Results on 3DMatch and 3DLoMatch}
We further detail the performance of \OURS{} with scene-wise results and 2 more metrics~(\textbf{RRE} and \textbf{RTE}) in Tab.~\ref{tab:scene_detailed}. The results further show the superiority of \OURS{} in scene-level registration.

\begin{table*}[h]
\caption{\textbf{Scene-Wise Results on 3DMatch and 3DLoMatch with \#Samples=5,000.} Best
performance is highlighted in bold while the second best is marked with an underline.}
\resizebox{\textwidth}{!}{
\begin{tabular}{lccccccccc|ccccccccc}
\toprule
\multirow{2}{*}{\textBF{Method}} & & & &\multicolumn{3}{c}{3DMatch}& & & &  & & &\multicolumn{3}{c}{3DLoMatch}& & & \\
 &Kitchen &Home\_1 &Home\_2 &Hotel\_1 &Hotel\_2 &Hotel\_3 &Study &Lab &Mean  &Kitchen &Home\_1 &Home\_2 &Hotel\_1 &Hotel\_2 &Hotel\_3 &Study &Lab & Mean\\  
\hline
\hline
&\multicolumn{17}{c}{\textit{Registration Recall}(\%)$\uparrow$}\\
\hline
3DSN~\cite{gojcic2019perfect} &90.6 &90.6 &65.4 &89.6 &82.1 &80.8 &68.4 &60.0 &78.4 &51.4 &25.9 &44.1 &41.1 &30.7 &36.6 &14.0 &20.3 &33.0\\
FCGF~\cite{choy2019fully} &\textBF{98.0} &94.3 &68.6 &96.7 &91.0 &\underline{84.6} &76.1 &71.1 &85.1 &60.8 &42.2 &53.6 &53.1 &38.0 &26.8 &16.1 &30.4 &40.1\\
D3Feat~\cite{bai2020d3feat} &96.0 &86.8 &67.3 &90.7 &88.5 &80.8 &78.2 &64.4 &81.6 &49.7 &37.2 &47.3 &47.8 &36.5 &31.7 &15.7 &31.9 &59.8\\
Predator~\cite{huang2021predator} &97.6 &\underline{97.2} &\underline{74.8} &\textBF{98.9} &\textBF{96.2} &\textBF{88.5} &\underline{85.9} &73.3 &89.0 &71.5 &58.2 &60.8 &77.5 &\underline{64.2} &61.0 &45.8 &39.1 &59.8\\
CoFiNet~\cite{yu2021cofinet} &96.4 &\textBF{99.1} &73.6 &95.6 &91.0 &\underline{84.6} &\textBF{89.7} &\textBF{84.4} &\textBF{89.3} &\underline{76.7} &\textBF{66.7} &\textBF{64.0} &\textBF{81.3} &\textBF{65.0} &\textBF{63.4} &\textBF{53.4} &\textBF{69.6} &\textBF{67.5}\\
\OURS{}~&\underline{97.8} &93.4 &\textBF{76.7} &\underline{98.4} &\underline{93.6} &\underline{84.6} &\underline{85.9} &\textBF{84.4} &\textBF{89.3} &\textBF{77.8} &\underline{60.6} &\underline{63.5} &\underline{79.4} &62.0 &\textBF{63.4} &\underline{48.7} &\underline{65.2} &\underline{65.1}\\
\hline
&\multicolumn{17}{c}{\textit{Relative Rotation Error}(\degree)$\downarrow$}\\
\hline
3DSN~\cite{gojcic2019perfect} &1.926 &1.843 &2.324 &2.041 &1.952 &2.908 &2.296 &2.301 &2.199 &3.020 &3.898 &3.427 &3.196 &3.217 &3.328 &4.325 &3.814 &3.528\\
FCGF~\cite{choy2019fully} &\textBF{1.767} &1.849 &\underline{2.210} &1.867 &1.667 &2.417 &\underline{2.024} &\textBF{1.792} &\underline{1.949} &\textBF{2.904} &3.229 &\underline{3.277} &2.768 &\textBF{2.801} &\underline{2.822} &\underline{3.372} &4.006 &3.147\\
D3Feat~\cite{bai2020d3feat} &2.016 &2.029 &2.425 &1.990 &1.967 &2.400 &2.346 &2.115 &2.161 &3.226 &3.492 &3.373 &3.330 &3.165 &2.972 &3.708 &3.619 &3.361\\
Predator~\cite{huang2021predator} &1.861 &\underline{1.806} &2.473 &2.045 &\underline{1.600} &2.458 &2.067 &\underline{1.926} &2.029 &3.079 &\textBF{2.637} &\textBF{3.220} &2.694 &2.907 &3.390 &\textBF{3.046} &3.412 &\underline{3.048}\\
CoFiNet~\cite{yu2021cofinet} &1.910 &1.835 &2.316 &\underline{1.767} &1.753 &\textBF{1.639} &2.527 &2.345 &2.011 &3.213 &3.119 &3.711 &2.842 &\underline{2.897} &3.194 &4.126 &\underline{3.138} &3.280\\
\OURS{}~&\underline{1.789} &\textBF{1.538} &\textBF{1.981} &\textBF{1.677} &\textBF{1.598} &\underline{1.935} &\textBF{1.833} &2.033 &\textBF{1.798} &\underline{2.987} &\underline{2.722} &3.313 &2.743 &2.956 &\textBF{2.439} &3.836 &\textBF{3.135} &\textBF{3.016}\\
\hline
&\multicolumn{17}{c}{\textit{Relative Translation Error}(m)$\downarrow$}\\
\hline
3DSN~\cite{gojcic2019perfect} &0.059 &0.070 &0.079 &0.065 &0.074 &0.062 &0.093 &0.065 &0.071 &0.082 &0.098 &0.096 &0.101 &\textBF{0.080} &0.089 &0.158 &\underline{0.120} &0.103\\
FCGF~\cite{choy2019fully} &0.053 &0.056 &0.071 &\underline{0.062} &0.061 &0.055 &0.082 &0.090 &0.066 &0.084 &0.097 &\textBF{0.076} &0.101 &0.084 &0.077 &0.144 &0.140 &0.100\\
D3Feat~\cite{bai2020d3feat} &0.053 &0.065 &0.080 &0.064 &0.078 &0.049 &0.083 &\underline{0.064} &0.067 &0.088 &0.101 &0.086 &\underline{0.099} &0.092 &\underline{0.075} &0.146 &0.135 &0.103\\
Predator~\cite{huang2021predator} &0.048 &\underline{0.055} &0.070 &0.073 &0.060 &0.065 &\underline{0.080} &\textBF{0.063} &0.064 &0.081 &\underline{0.080} &0.084 &\underline{0.099} &0.096 &0.077 &\textBF{0.101} &0.130 &\underline{0.093}\\
CoFiNet~\cite{yu2021cofinet} &\underline{0.047} &0.059 &\underline{0.063} &0.063 &\textBF{0.058} &\underline{0.044} &0.087 &0.075 &\underline{0.062} &\underline{0.080} &\textBF{0.078} &\underline{0.078} &\underline{0.099} &0.086 &0.077 &0.131 &0.123 &0.094\\
\OURS{}~&\textBF{0.044} &\textBF{0.048} &\textBF{0.056} &\textBF{0.060} &\underline{0.059} &\textBF{0.040} &\textBF{0.071} &0.071 &\textBF{0.056} &\textBF{0.078} &0.082 &0.085 &\textBF{0.094} &\underline{0.082} &\textBF{0.059} &\underline{0.116} &\textBF{0.114} &\textBF{0.089}\\
\bottomrule
\end{tabular}}
\label{tab:scene_detailed}
\end{table*}

\begin{table*}[h!]
\caption{\textbf{Ablation Study on 3DLoMatch and Rotated 3DLoMatch with \# Samples = 5,000.} In the brackets are the changes compared to baseline \OURS{}.}
\resizebox{\textwidth}{!}{
\begin{tabular}{llccc|ccc}
\toprule
\multirow{2}{*}{\textBF{Ablation Part}} &\multirow{2}{*}{\textBF{Models}} &\multicolumn{3}{c}{\textBF{3DLoMatch}}  &\multicolumn{3}{c}{\textBF{3DLoMatch~(Rotated)}}\\
& &\textBF{IR}(\%) $\uparrow$ &\textBF{FMR}(\%) $\uparrow$ &\textBF{RR}(\%) $\uparrow$  &\textBF{IR}(\%) $\uparrow$ &\textBF{FMR}(\%) $\uparrow$ &\textBF{RR}(\%)$\uparrow$\\
\hline
\hline
(0)~\textit{None} &\OURS{}~(\textit{Baseline}) &32.1 &85.1 &65.1 &32.1 &84.5 &66.9
\\
\hline
\multirow{4}{*}{(1)~\textit{Local Description}} &(a)~xyz &20.8{(\textcolor{red}{$-$11.3})} &77.5{(\textcolor{red}{$-$7.60})} &56.0{(\textcolor{red}{$-$9.10})} &20.2{(\textcolor{red}{$-$11.9})} &76.2{(\textcolor{red}{$-$8.30})} &57.4{(\textcolor{red}{$-$9.50})}
\\
&(b)~relative xyz & 25.7{(\textcolor{red}{$-$6.40})} &79.6{(\textcolor{red}{$-$5.50})} &58.5{(\textcolor{red}{$-$6.60})} &24.9{(\textcolor{red}{$-$7.20})} &79.9{(\textcolor{red}{$-$4.60})} &59.9{(\textcolor{red}{$-$7.00})}
\\
&(c)~xyz + PPF &31.1{(\textcolor{red}{$-$1.00})} &85.1{(+0.00)} &65.3{(\textcolor{blue}{+0.20})} &31.1{(\textcolor{red}{$-$1.00})} &83.6{(\textcolor{red}{$-$0.90})} &66.5{(\textcolor{red}{$-$0.40})}
\\
&(d)~relative xyz + PPF &30.7{(\textcolor{red}{$-$1.40})} &83.6{(\textcolor{red}{$-$1.50})} &62.5{(\textcolor{red}{$-$2.60})} &30.6{(\textcolor{red}{$-$1.50})} &83.1{(\textcolor{red}{$-$1.40})} &64.5{(\textcolor{red}{$-$2.40})}
\\
\hline
\multirow{4}{*}{(2)~\textit{Global Description}} &(a)~none &13.8{(\textcolor{red}{$-$18.3})} &75.4{(\textcolor{red}{$-$9.70})} &61.1{(\textcolor{red}{$-$4.00})} &13.9{(\textcolor{red}{$-$18.2})} &76.0{(\textcolor{red}{$-$8.50})} &66.0{(\textcolor{red}{$-$0.90})}
\\
&(b)~xyz &18.6{(\textcolor{red}{$-$13.5})} &81.3{(\textcolor{red}{$-$3.80})} &65.1{(+0.00)} &18.5{(\textcolor{red}{$-$13.6})} &80.5{(\textcolor{red}{$-$4.00})} &65.8{(\textcolor{red}{$-$1.10})}
\\
&(c)~relative xyz &15.1{(\textcolor{red}{$-$17.0})} &77.5{(\textcolor{red}{$-$7.60})} &62.8{(\textcolor{red}{$-$2.30})} &14.8{(\textcolor{red}{$-$17.3})} &75.7{(\textcolor{red}{$-$8.80})} &65.2{(\textcolor{red}{$-$1.70})}
\\
&(d)~xyz+sinusoidal~\cite{vaswani2017attention} &15.1(\textcolor{red}{$-$17.0}) &76.7(\textcolor{red}{$-$8.40}) &64.6(\textcolor{red}{$-$0.50}) &15.1(\textcolor{red}{$-$17.0}) &78.3(\textcolor{red}{$-$6.20}) &66.5(\textcolor{red}{$-$0.40})
\\
\hline
\multirow{4}{*}{(3)~\textit{Attention Blocks}} &(a)~\textit{K}=0 &10.2{(\textcolor{red}{$-$21.9})} &60.8{(\textcolor{red}{$-$24.3})} &50.0{(\textcolor{red}{$-$15.1})} &10.3{(\textcolor{red}{$-$21.8})} &60.2{(\textcolor{red}{$-$24.3})} &53.6{(\textcolor{red}{$-$13.3})}
\\
&(b)~\textit{K}=1 &16.6{(\textcolor{red}{$-$15.5})} &77.1{(\textcolor{red}{$-$8.00})} &63.0{(\textcolor{red}{$-$2.10})} &16.6{(\textcolor{red}{$-$15.5})} &77.8{(\textcolor{red}{$-$6.70})} &66.1{(\textcolor{red}{$-$0.80})}
\\
&(c)~\textit{K}=3 &24.9{(\textcolor{red}{$-$7.20})} &82.2{(\textcolor{red}{$-$2.90})} &65.1{(+0.00)} &25.0{(\textcolor{red}{$-$7.10})} &82.4{(\textcolor{red}{$-$2.10})} &66.6{(\textcolor{red}{$-$0.30})}
\\
&(d)~\textit{K}=10 &32.5{(\textcolor{blue}{+0.40})} &83.6{(\textcolor{red}{$-$1.50})} &63.6{(\textcolor{red}{$-$1.50})} &32.4{(\textcolor{blue}{+0.30})} &83.4{(\textcolor{red}{$-$1.10})} &66.8{(\textcolor{red}{$-$0.10})}
\\
\bottomrule
\end{tabular}}
\label{tab:ablation_3dlomatch}
\end{table*}

\subsection{Ablation Study}
We ablate different parts of \OURS{}, including (1)~\textit{Local Description}, (2)~\textit{Global Description} and (3)~\textit{Attention Blocks} to assess the importance of each individual component. We use 3DMatch and 3DLoMatch, together with their rotated versions for ablation study. Detailed results are found in Tab.~\ref{tab:ablation_3dlomatch} for 3DLoMatch and Rotated 3DLoMatch, and in the Appendix for 3DMatch and Rotated 3DMatch.

\subsubsection{Local Description} 
In the ablation of (1)~\textit{Local Description}, we replace our local PPF-based geometric description with two rotation-variant variants: (a)~xyz - learning local descriptors from the raw 3D coordinates of all the points in the support area around each node; and (b)~relative xyz - learning descriptors from relative 3D coordinates of points w.r.t. the central node of the support area. In both cases, the performance drops compared to the baseline \OURS{}, which indicates the power of our PPF signature-based geometric description. Moreover, we observe a more significant drop in performance in terms of \textbf{IR} and \textbf{FMR} when facing larger rotations which further demonstrates the importance of rotational invariance. Similarly to~\cite{deng2018ppfnet,yew2020rpm}, we also concatenate PPF signatures with coordinates of points for local description in (c) and (d). This results in a better performance than the variants with only 3D coordinates but still perform slightly worse than the baseline \OURS{}. Thanks to the global awareness in \OURS{}, it is unnecessary to supplement PPF with global coordinates, as in (c), to incorporate global contexts. Pure local geometry which is rotation-invariant already promises good performance.

\subsubsection{Global Description}

We first ablate (2)~\textit{Global Description} by removing structural descriptors learned from our proposed global PPF signatures. As shown in (a), this significantly damages the performance especially in terms of \textbf{IR}, which proves the importance of informing local descriptors with global structural cues. 
To further prove the significance of our rotation-invariant structural description, we replace the structural descriptors in baseline \OURS{} with (b)xyz - learning global positional descriptors from the raw 3D coordinates of each node, and (c) ~relative xyz - learning global positional descriptors from the relative position of each node w.r.t. the other nodes in the same frame. Moreover, we also follow~\cite{vaswani2017attention} to learn descriptors from node coordinates projected by sinusoidal functions~\cite{van2008visualizing} in (d). The decreased performance of all the variants further confirms the superiority of our design of encoding structural descriptors from global PPF signatures.

\subsubsection{Attention Blocks}
To emphasize the importance of global awareness, we ablate \OURS{} with different number of (3)~\textit{Attention Blocks}. In (a), we remove all the attention blocks (\textit{K}=0) and only use the globally-informed descriptors, which leads to a sharp decrease of the performance. This proves the significance of global awareness obtained from learned global contexts. When we increase the number of attention blocks to (b)~\textit{K}=1 and (c)~\textit{K}=3, the performance increases correspondingly, though it does not reach the baseline performance with \textit{K}=6. This observation indicates that stronger global awareness improves the overall performance. However, when we keep including more and more Attention Blocks in (d)~\textit{K}=10, the performance only stays on-par with \OURS{} baseline, indicating that using 6 Attention Blocks is a proper option with good performance.



%% file: sections/5conclusion.tex
\section{Conclusion}
In this paper, we introduce \OURS{} with a ViT architecture that learns both rotation-invariant and globally-aware descriptors, upon which correspondences are established in a coarse-to-fine manner for point cloud registration. We learn from rotation-invariant PPFs for encoding local geometry and further introduce global PPF signatures to encode a node-specific structural description of the whole scene. The structural descriptors learned from global PPF signatures strengthen local descriptors with the global 3D structures in a rotation-invariant fashion. The distinctiveness of descriptors is further enhanced in the consecutive attention blocks with the learned geometric context across the whole scene. The coarse-to-fine mechanism is further leveraged to establish reliable correspondences upon our powerful \OURS{} descriptors. Experimental results confirm the effectiveness of our approach on both object and scene-level data. We hope our work can inspire more research looking toward the joint rotational invariance and distinctiveness of descriptors in point cloud registration.

%% file: sections/6appendix.tex
\section{Appendix}
In this Appendix, we first detail related metrics in Sec.~\ref{sec:metrics}. We then demonstrate the inference speed of \OURS{} in Sec.~\ref{sec:runtime} and detail the ablation study on 3DMatch~\cite{zeng20173dmatch} and Rotated 3DMatch in Sec.~\ref{sec:more_ablation}. We further use KITTI~\cite{geiger2012we} to demonstrate our robustness against poor normal estimation in Sec.~\ref{sec:kitti}. Finally, more quantitative results are illustrated in Sec.~\ref{sec:more_qualitative}.

\subsection{Detailed Metrics}
\label{sec:metrics}
\noindent \textbf{Relative Rotation and Translation Errors.}
Given the estimated rotation $\mathbf{\overline{R}}\in SO(3)$ and translation $\mathbf{\overline{t}}\in \mathbb{R}^3$ between a pair of point clouds $(\mathcal{X}, \mathcal{Y})$, the \textit{Relative Rotation Error} and the \textit{Relative Translation Error} w.r.t. the ground truth rotation $\mathbf{R}\in SO(3)$ and translation $\mathbf{t}\in \mathbb{R}^3$ are computed as:

\begin{equation}
  \begin{aligned}
\mathbf{RRE}&(\mathcal{X}, \mathcal{Y}) = arccos(\frac{trace(\mathbf{R}^T\overline{\mathbf{R}}) - 1}{2}), and \\
\mathbf{RTE}&(\mathcal{X}, \mathcal{Y}) = \lVert \mathbf{t} - \mathbf{\overline{t}}\rVert_2,
\end{aligned}
\label{eq: rre_rte}
\end{equation}
\noindent respectively.

\noindent \textbf{Root-Mean-Square Error.}
Given the estimated transformation $\overline{\mathbf{T}}\in SE(3)$ and the ground truth transformation $\mathbf{T}$ between a pair of point clouds $(\mathcal{X}, \mathcal{Y})$, there can be two ways to calculate \textit{Root-Mean-Square Error}~(\textbf{RMSE}). According to~\cite{yuan2020deepgmr}, the first way to calculate \textbf{RMSE} reads as:
\begin{equation}
    \mathbf{RMSE_1}(\mathcal{X}, \mathcal{Y}) = \frac{1}{|\mathcal{X}|}\sqrt{\sum_{\mathbf{x}\in \mathcal{X}}\lVert\overline{\mathbf{T}}(\mathbf{x}) - \mathbf{T}(\mathbf{x})\rVert_2^2},
    \label{eq:rmse_1}
\end{equation}
which is used for the experiments on ModelNet~\cite{wu20153d} and the calculation of \textit{Registration Recall} on rotated 3DMatch~\cite{zeng20173dmatch} and rotated 3DLoMatch~\cite{huang2021predator}. Additionally, we follow~\cite{huang2021predator} to calculate $\mathbf{RMSE_2}$, upon which the \textit{Registration Recall} on 3DMatch and 3DLoMatch is further defined. $\mathbf{RMSE_2}$ is calculated as:

\begin{equation}
    \mathbf{RMSE_2}(\mathcal{X}, \mathcal{Y}) = \sqrt{\frac{1}{|\mathcal{C}^*|}\sum_{(\mathbf{x}, \mathbf{y})\in \mathcal{C}^*} \lVert\overline{\mathbf{T}}(\mathbf{x}) - \mathbf{y}\rVert_2^2},
    \label{eq:rmse_2}
\end{equation}
\noindent where $\mathcal{C}^*$ is a ground truth correspondence set.

\noindent \textbf{Inlier Ratio.}
\textit{Inlier Ration}~(\textbf{IR}) measures the fraction of putative correspondences $(\mathbf{x}, \mathbf{y})\in \hat{\mathcal{C}}$ s.t. $\lVert\mathbf{T}(\mathbf{x}) - \mathbf{y}\rVert_2$ is within a threshold $\tau_1=10cm$, where $\mathbf{T}$ stands for the ground truth transformation between point clouds $\mathcal{X}$ and $\mathcal{Y}$. The $\mathbf{IR}$ of a single point cloud pair $(\mathcal{X}, \mathcal{Y})$ with a putative correspondence set $\hat{\mathcal{C}}$ is defined as:
\begin{equation}
\mathbf{IR}(\mathcal{X}, \mathcal{Y}) = \frac{1}{|\hat{\mathcal{C}}|} \sum_{(\mathbf{x}, \mathbf{y})\in \hat{\mathcal{C}}} \mathds{1}(\lVert\mathbf{T}(\mathbf{x}) - \mathbf{y}\rVert_2 < \tau_1),
\label{eq:ir}    
\end{equation}
\noindent where $\mathds{1}(\cdot)$ denotes the indicator function.

\noindent\textbf{Feature Matching Recall.}
\textit{Feature Matching Recall}~(\textbf{FMR}) counts the fraction of point cloud pairs $(\mathcal{X}, \mathcal{Y})$ that satisfies $\mathbf{IR}(\mathcal{X}, \mathcal{Y}) > \tau_2$, which is set to 5\% in our experiments. Given a dataset $\mathcal{H}$ consisting of $|\mathcal{H}|$ point cloud pairs, the \textbf{FMR} is computed as:
\begin{equation}
\mathbf{FMR}(\mathcal{H}) = \frac{1}{|\mathcal{H}|} \sum_{(\mathcal{X}, \mathcal{Y})\in\mathcal{H}} \mathds{1}(\mathbf{IR}(\mathcal{X}, \mathcal{Y}) > \tau_2).
\label{eq:fmr}
\end{equation}

\begin{table}
\centering
\caption{\textbf{Runtime.} All the reported time is averaged over the whole 3DMatch testing set, which consists of 1,623 point cloud pairs. "Desc" reports the runtime for description, i.e., from data loading to the generation of descriptors. "Reg" reports the time for registration, i.e., from the generated descriptors to the estimation of rigid transformation via RANSAC~\cite{fischler1981random}. These two parts of time sum to "Total".}
\begin{tabular}{lccccc}
\toprule
Method &Desc~(s)$\downarrow$ &Reg~(s)$\downarrow$ &Total~(s)$\downarrow$\\
\hline
\hline
SpinNet~\cite{ao2021spinnet} &44.92 &- &\textgreater 44.92\\
Predator~\cite{huang2021predator}  &\underline{0.506} &0.677 &1.183\\
CoFiNet~\cite{yu2021cofinet}  &\textBF{0.145} &\textBF{0.043} &\textBF{0.188}\\
\OURS{}~(\textit{Ours}) &0.731 &\underline{0.101} &\underline{0.832}\\
\bottomrule
\end{tabular}
\label{tab:runtime}
\end{table}

\begin{figure*}[h!]
\centering
\includegraphics[width=\linewidth]{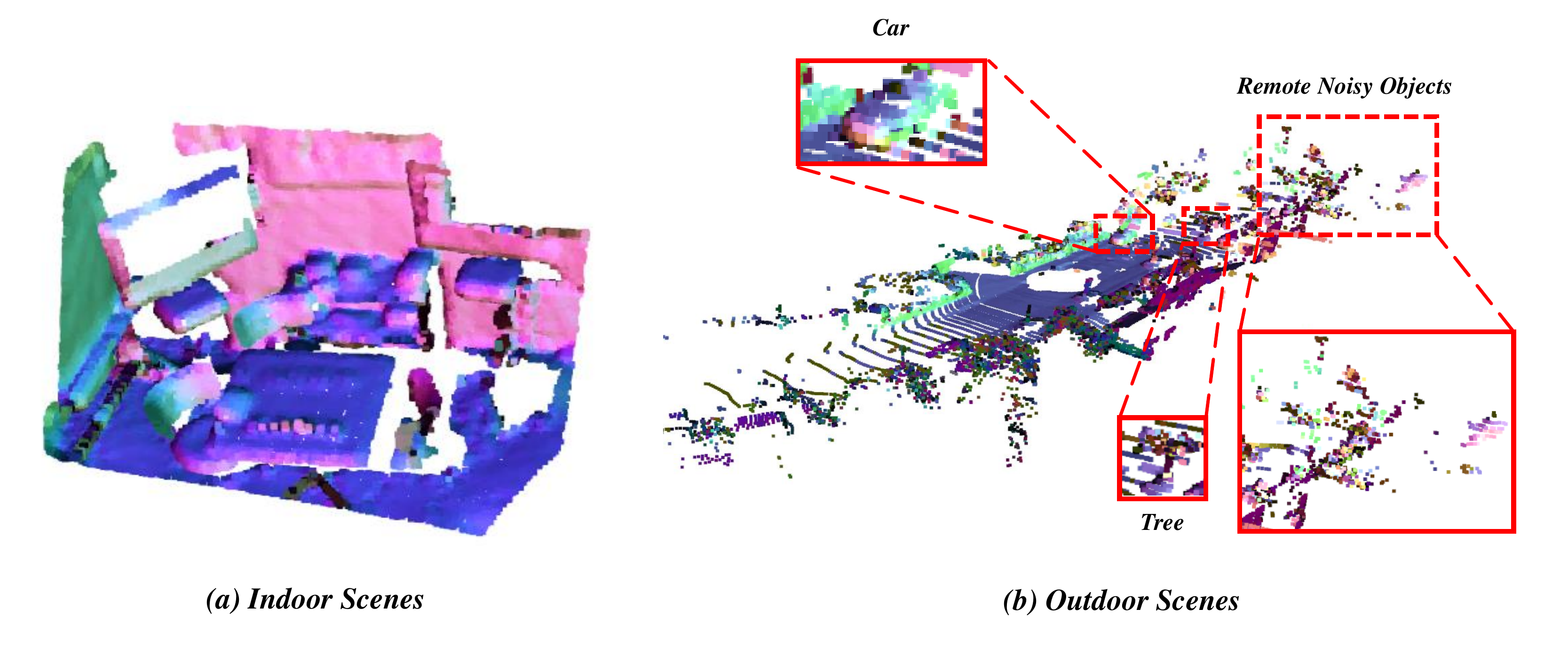}
\caption{\textbf{Demonstration of the Quality of Normal Estimation.} Normals are estimated by using Open3D~\cite{zhou2018open3d} and are color-coded for visualization. The indoor scene in column~(a) is from 3DMatch~\cite{zeng20173dmatch}, while the outdoor scene in column~(b) is from KITTI~\cite{geiger2012we}. For indoor scenes, the normal estimation is accurate, i.e., the colors are smooth in the visualization. However, the quality of estimated normals in outdoor scenes is much worse. Although the estimated normals are not bad for the "Car" which is represented clearly by points with less noise, the normals of the "Tree" are worse, due to its complex geometry and noisy representation. Moreover, the objects that are far away from the LiDAR and roughly represented by sparse points are hard to recognize and with the worst normal quality. }
\label{fig:normal}
\end{figure*}

\begin{table*}[h]
\caption{\textbf{Ablation study on 3DMatch and Rotated 3DMatch.} In the brackets are the changes compared to baseline \OURS{}. \# Samples = 5,000.}
\resizebox{1.0\textwidth}{!}{
\begin{tabular}{llccc|ccc}
\toprule
\multirow{2}{*}{\textBF{Ablation Part}} &\multirow{2}{*}{\textBF{Models}} &\multicolumn{3}{c}{\textBF{3DMatch}}  &\multicolumn{3}{c}{\textBF{3DMatch~(Rotated)}}  \\
& &\textBF{IR}(\%) $\uparrow$ &\textBF{FMR}(\%) $\uparrow$ &\textBF{RR}(\%) $\uparrow$  &\textBF{IR}(\%) $\uparrow$ &\textBF{FMR}(\%) $\uparrow$ &\textBF{RR}(\%) $\uparrow$\\
\hline
\hline
(0)~\textit{None} &\OURS{}~(\textit{Baseline}) &68.4 &97.9 &89.3 &68.5 &98.2 &93.0
\\
\hline
\multirow{4}{*}{(1)~\textit{Local Description}} &(a)~xyz &53.7(\textcolor{red}{$-$14.7}) &96.1(\textcolor{red}{$-$1.80}) &86.8(\textcolor{red}{$-$2.50}) &52.7(\textcolor{red}{$-$15.8}) &95.8(\textcolor{red}{$-$2.40}) &89.1(\textcolor{red}{$-$3.10})
\\
&(b)~relative xyz &60.9(\textcolor{red}{$-$7.50}) &97.2(\textcolor{red}{$-$0.70}) &87.5(\textcolor{red}{$-$1.80}) &60.0(\textcolor{red}{$-$8.50}) &96.4(\textcolor{red}{$-$1.80}) &90.3(\textcolor{red}{$-$2.70})
\\
&(c)~xyz + PPF &66.3(\textcolor{red}{$-$2.10}) &98.2(\textcolor{blue}{+0.30}) &88.5(\textcolor{red}{$-$0.80}) &65.9(\textcolor{red}{$-$2.60}) &98.1(\textcolor{red}{$-$0.10}) &92.4(\textcolor{red}{$-$0.60})
\\
&(d)~relative xyz + PPF &66.8(\textcolor{red}{$-$1.60}) &97.5(\textcolor{red}{$-$0.40}) &87.7(\textcolor{red}{$-$1.60}) &66.7(\textcolor{red}{$-$1.80}) &97.4(\textcolor{red}{$-$0.80}) &92.1(\textcolor{red}{$-$0.90})
\\
\hline
\multirow{4}{*}{(2)~\textit{Global Description}} &(a)~none &34.9(\textcolor{red}{$-$33.5}) &97.0(\textcolor{red}{$-$0.90}) &88.1(\textcolor{red}{$-$1.20}) &35.0(\textcolor{red}{$-$33.5}) &97.0(\textcolor{red}{$-$1.20}) &92.8(\textcolor{red}{$-$0.20})
\\
&(b)~xyz &42.3(\textcolor{red}{$-$26.1}) &97.8(\textcolor{red}{$-$0.10}) &87.7(\textcolor{red}{$-$1.60}) &42.3(\textcolor{red}{$-$26.2}) &97.6(\textcolor{red}{$-$0.60}) &92.3(\textcolor{red}{$-$0.70})
\\
&(c)~relative xyz &37.2(\textcolor{red}{$-$31.2}) &97.0(\textcolor{red}{$-$0.90}) &88.0(\textcolor{red}{$-$1.30}) &37.0(\textcolor{red}{$-$31.5}) &96.8(\textcolor{red}{$-$1.40}) &93.3(\textcolor{blue}{+0.30})
\\
&(d)~xyz+sinusoidal~\cite{vaswani2017attention} &37.1(\textcolor{red}{$-$31.3}) &97.1(\textcolor{red}{$-$0.80}) &89.8(\textcolor{blue}{+0.50}) &37.1(\textcolor{red}{$-$31.4}) &97.6(\textcolor{red}{$-$0.60}) &93.0(+0.00)
\\
\hline
\multirow{4}{*}{(3)~\textit{Attention Blocks}} &(a)~\textit{K}=0 &29.7(\textcolor{red}{$-$38.7}) &94.2(\textcolor{red}{$-$3.70}) &83.0(\textcolor{red}{$-$6.30}) &29.5(\textcolor{red}{$-$39.0}) &94.2(\textcolor{red}{$-$4.00}) &89.3(\textcolor{red}{$-$3.70})
\\
&(b)~\textit{K}=1 &43.7(\textcolor{red}{$-$24.7}) &97.4(\textcolor{red}{$-$0.50}) &90.1(\textcolor{blue}{+0.80}) &43.7(\textcolor{red}{$-$24.8}) &97.3(\textcolor{red}{$-$0.90}) &93.4(\textcolor{blue}{+0.40})
\\
&(c)~\textit{K}=3 &58.1(\textcolor{red}{$-$10.3}) &97.7(\textcolor{red}{$-$0.20}) &88.8(\textcolor{red}{$-$0.50}) &58.4(\textcolor{red}{$-$10.1}) &98.1(\textcolor{red}{$-$0.10}) &92.7(\textcolor{red}{$-$0.30})
\\
&(d)~\textit{K}=10 &68.5(\textcolor{blue}{+0.10}) &98.1(\textcolor{blue}{+0.20}) &89.0(\textcolor{red}{$-$0.30}) &68.4(\textcolor{red}{$-$0.10}) &97.9(\textcolor{red}{$-$0.30}) &92.2(\textcolor{red}{$-$0.80})
\\
\bottomrule
\end{tabular}}
\label{tab:ablation_3dmatch}
\end{table*}

 \begin{table}[h]
\caption{\textbf{Quantitative comparisons on KITTI.} Best performance is highlighted in bold.}
\centering
\resizebox{0.5\textwidth}{!}{
\begin{tabular}{l|ccc}
\toprule
 Method &RTE(cm)$\downarrow$ &RRE($^{\circ}$)$\downarrow$ &RR(\%)$\uparrow$\\
\hline
\hline
3DFeat-Net~\cite{yew20183dfeat} &25.9 &0.57 &96.0\\
FCGF~\cite{choy2019fully} &9.5 &0.30 &96.6\\
D3Feat~\cite{bai2020d3feat} &7.2 &0.30 &\textbf{99.8} \\
SpinNet~\cite{ao2021spinnet} &9.9 &0.47 &99.1\\
Predator~\cite{huang2021predator} &\textbf{6.8} &\textbf{0.27} &\textbf{99.8}\\
CoFiNet~\cite{yu2021cofinet} &8.5 &0.41 &\textbf{99.8}\\
\OURS{}~(\textit{Ours}) &13.5 &0.45 &99.1\\
\bottomrule
\end{tabular}}
\label{tab:kitti}
\end{table}

\noindent \textbf{Registration Recall.}
\textit{Registration Recall}~(\textbf{RR}) that measures the fraction of successfully registered point cloud pairs directly evaluates the performance of a method on the task of point cloud registration. More specifically, it counts the fraction of point cloud pairs $(\mathcal{X}, \mathcal{Y})$ that satisfies $\mathbf{RMSE_2}(\mathcal{X}, \mathcal{Y}) < \tau_3$, where $\tau_3$ is set to 0.2m in our experiments. Given a dataset with $|\mathcal{H}|$ point cloud pairs, \textbf{RR} is defined as:
\begin{equation}
\mathbf{RR}(\mathcal{H}) = \frac{1}{|\mathcal{H}|} \sum_{(\mathcal{X}, \mathcal{Y})\in \mathcal{H}} \mathds{1}(\mathbf{RMSE_2}(\mathcal{X}, \mathcal{Y}) < \tau_3).
\label{eq:rr}
\end{equation}

\subsection{Runtime Analysis}
\label{sec:runtime}
We test all the following approaches on a machine with "AMD Ryzen 7 5800X @ 3.80GHZ $\times$ 8" CPU and "NVIDIA GeForce RTX 3090" GPU. In Tab.~\ref{tab:runtime} we compare \OURS{} with 3 state-of-the-art methods in terms of runtime. Among all the baselines, SpinNet~\cite{ao2021spinnet} is a patch-based rotation-invariant method, while Predator~\cite{huang2021predator} and CoFiNet~\cite{yu2021cofinet} are globally-aware models with fully-convolutional encoder-decoder architectures. As \OURS{} use a ViT architecture that starts from the description of local regions, when compared to Predator and CoFiNet, it takes more time to generate descriptors. However, \OURS{} generates descriptors much faster than SpinNet, as our global awareness simplifies the feature engineering on local regions, and our mechanism in tackling the repeatability issues significantly reduces the number of required local regions. For registration time, as we adopt a coarse-to-fine strategy, the runtime is significantly reduced when compared to Predator. Moreover, we use the second least total time among all the methods, which demonstrates our efficiency for the task of point cloud registration.

\subsection{Ablation Studies on 3DMatch/Rotated 3DMatch}
\label{sec:more_ablation}
In Tab.~\ref{tab:ablation_3dmatch}, we show the ablation study on 3DMatch and rotated 3DMatch with the same setting as in the ablation study of 3DLoMatch and rotated 3DLoMatch in the main paper. Different variants behave similarly , which further illustrates the significance of each individual part of \OURS{}.

\subsection{Robustness against Poor Normal Estimation}
\label{sec:kitti}
As our inherent rotational invariance is affected by the quality of the estimated normals, we further conduct extensive experiments on KITTI~\cite{geiger2012we} which consists of outdoor scans from LiDAR to prove the robustness of our \OURS{} descriptors against poor normal estimation. The estimated normals of both indoor and outdoor scenarios are visualized in Fig.~\ref{fig:normal} to show the poor normal estimation for outdoor scenes compared to indoor ones. Under this circumstance, as shown in Tab.~\ref{tab:kitti}, although \OURS{} is affected by the poor normal quality, it still performs on par with those state-of-the-art methods in terms of three different metrics.

\subsection{More Qualitative Results. }
\label{sec:more_qualitative}
\begin{figure*}[h!]
\centering
\includegraphics[width=\linewidth]{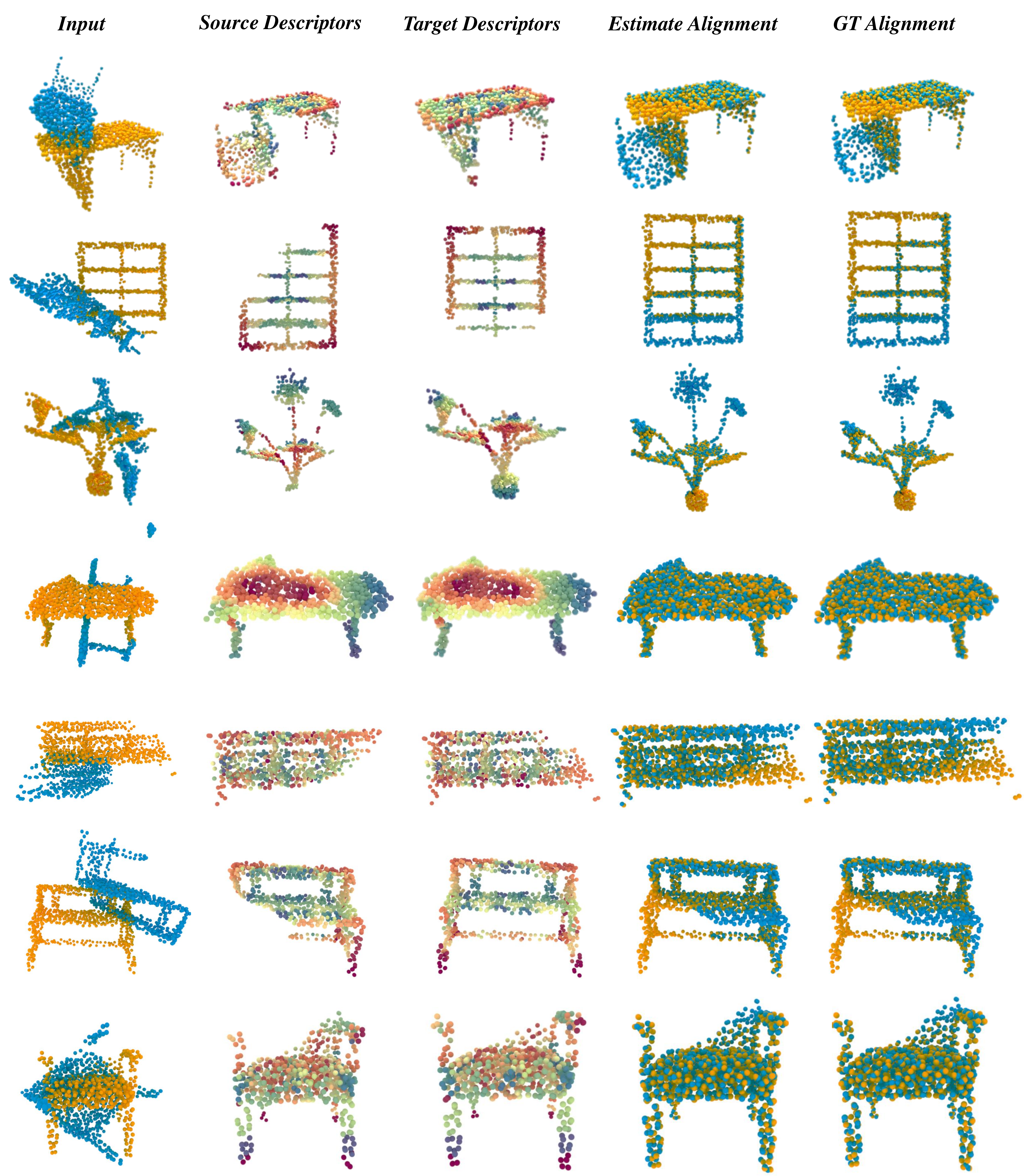}
\caption{\textbf{More Qalitative Results on ModelNet40.} We use t-SNE~\cite{van2008visualizing} to visualize the learned descriptors of source and target point clouds.}
\label{fig:more_modelnet}
\end{figure*}

\begin{figure*}[h!]
\centering
\includegraphics[width=\linewidth]{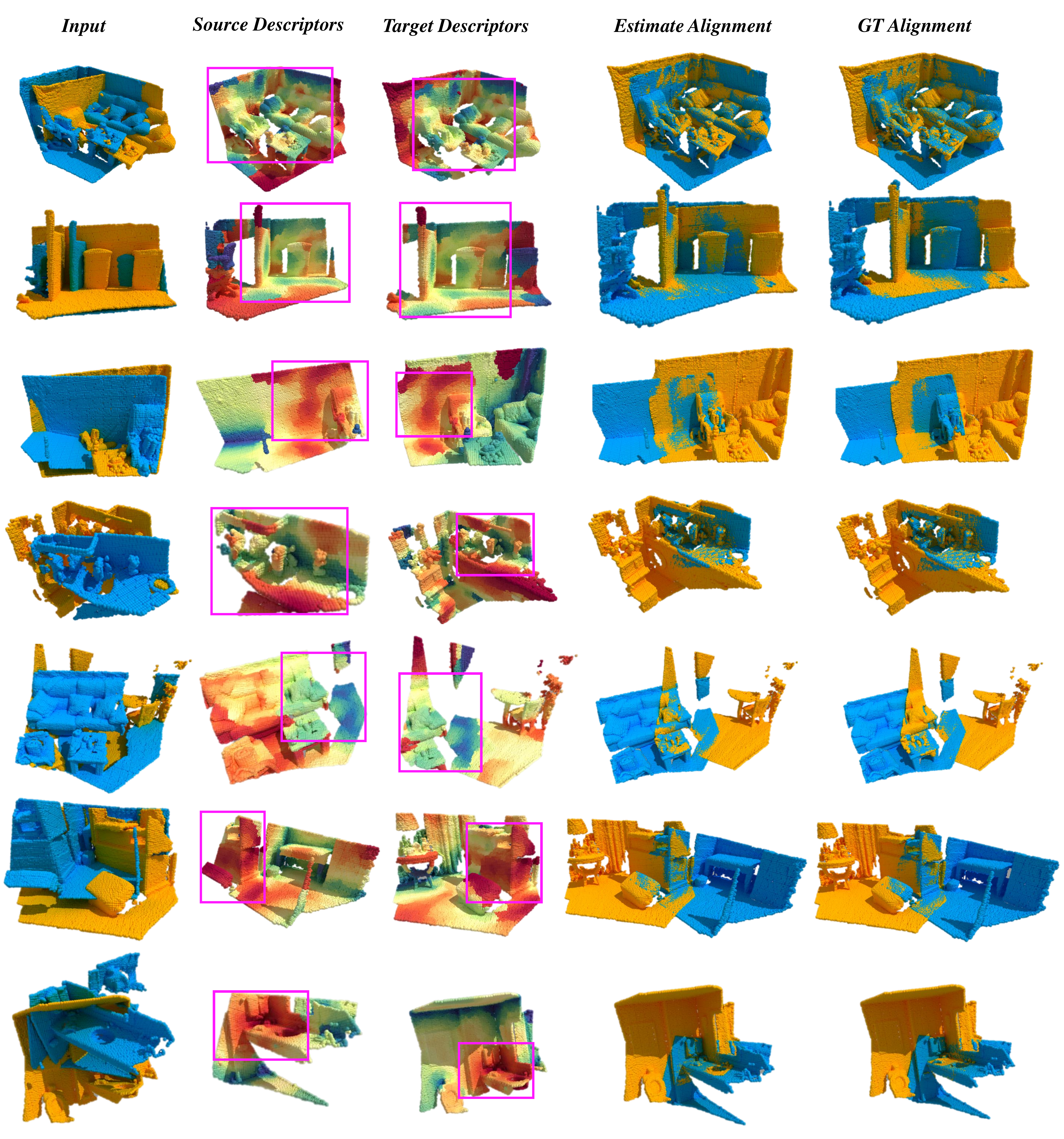}
\caption{\textbf{More qualitative results on 3DMatch/3DLoMatch.} We use t-SNE~\cite{van2008visualizing} to visualize the learned descriptors of source and target point clouds. In the rectangles, we roughly demonstrate the overlap regions.}
\label{fig:more_3dmatch}

\end{figure*}

More qualitative results on both ModetNet40 and 3DMatch/3DLoMatch can be found in Fig.~\ref{fig:more_modelnet} and Fig.~\ref{fig:more_3dmatch}, respectively. In each figure, the first column gives a pair of unaligned point clouds, where the source point cloud is presented as blue and the target point cloud is shown in yellow. The second and third columns illustrate the \OURS{} descriptors visualized by t-SNE~\cite{van2008visualizing} for source and target point clouds, respectively. The forth column demonstrates the estimated alignment, while the last column provides the ground truth one.

